\pdfoutput=1

\documentclass[11pt]{article} 
\usepackage[final]{acl}

\usepackage{amsmath}
\usepackage{amsfonts, bm}
\usepackage[T1]{fontenc}
\usepackage[pdftex]{graphicx}  
\usepackage{microtype}
\usepackage{hyperref}
\usepackage{url}
\usepackage{booktabs}
\usepackage{multirow}
\usepackage{subfig}
\usepackage{threeparttable}
\usepackage{tabularx}

\definecolor{darkblue}{rgb}{0, 0, 0.5}
\hypersetup{colorlinks=true, citecolor=darkblue, linkcolor=darkblue, urlcolor=darkblue}

\title{Parameter Efficient Reinforcement Learning from Human Feedback}

\author{Hakim Sidahmed\footnotemark[1]\textsuperscript{1}, Samrat Phatale\footnotemark[1]\textsuperscript{1}, Alex Hutcheson\textsuperscript{2}, Zhuonan Lin\textsuperscript{2}, \\ 
\bf Zhang Chen\textsuperscript{2},
 Zac Yu\textsuperscript{2}, Jarvis Jin\textsuperscript{2}, Simral Chaudhary\textsuperscript{1}, Roman Komarytsia\textsuperscript{2}, \\ 
 \bf Christiane Ahlheim\textsuperscript{2},
 Yonghao Zhu\textsuperscript{2}, Bowen Li\textsuperscript{2}, Saravanan Ganesh\textsuperscript{2}, Bill Byrne\textsuperscript{2},  \\ 
 \bf Jessica Hoffmann\textsuperscript{1},
 Hassan Mansoor\textsuperscript{1}, Wei Li\textsuperscript{1}, Abhinav Rastogi\textsuperscript{1}, Lucas Dixon\textsuperscript{1} \\
\textsuperscript{1}Google DeepMind,
\textsuperscript{2}Google \\
\small{
    \texttt{\{hsidahmed, samratph, ldixon\}@google.com}
}
}

\begin{document}

\maketitle
\renewcommand{\thefootnote}{\fnsymbol{footnote}}
\footnotetext[1]{Equal Contribution.}

\begin{abstract}
\label{sec:abstract}
While Reinforcement Learning from Human Feedback (RLHF) effectively aligns pretrained Large Language and Vision-Language Models (LLMs, and VLMs) with human preferences, its computational cost and complexity hamper its wider adoption.
To alleviate some of the computational burden of fine-tuning, parameter efficient methods, like LoRA \citep{hu2021lora} were introduced.
In this work, we empirically evaluate the setup of \textbf{P}arameter \textbf{E}fficient \textbf{R}einforcement \textbf{L}earning from \textbf{H}uman \textbf{F}eedback (PE-RLHF) that leverages LoRA fine-tuning for Reward Modeling, and Reinforcement Learning.
We benchmark the PE-RLHF setup on six diverse datasets spanning summarization, harmless/helpful response generation, UI automation, and visual question answering in terms of effectiveness of the trained models, and the training resources required.
Our findings show, for the first time, that PE-RLHF achieves comparable performance to RLHF, while significantly reducing training time (up to 90\% faster for reward models, and 30\% faster for RL), and memory footprint (up to 50\% reduction for reward models, and 27\% for RL).
We provide comprehensive ablations across LoRA ranks, and model sizes for both reward modeling and reinforcement learning.
By mitigating the computational burden associated with RLHF, we push for a broader adoption of PE-RLHF as an alignment technique for LLMs and VLMs.

\end{abstract}

\section{Introduction}
Large Language and Vision-Language Models (LLMs, and VLMs) like GPT-4 \citep{openai2023gpt4} and Gemini \citep{geminiteam2023gemini,reid2024gemini} demonstrate remarkable performance across diverse tasks.
However, aligning these models with human preferences remains crucial for ensuring desirable behavior \citep{bommasani2022opportunities}.
This alignment improves instruction following \citep{ouyang2022training}, and facilitates optimizing for behaviors that lack a clear mathematical loss function, such as safety properties \citep{bai2022training,bai2022constitutional,glaese2022improving}, helpfulness \citep{bai2022training,glaese2022improving}, summarization characteristics \citep{stiennon2020learning}, and visual instructions \citep{sun2023aligning}.
Reinforcement Learning from Human Feedback (RLHF) has emerged as a prominent method for achieving this alignment.
It involves training a reward model (RM) on human feedback data, and subsequently using this RM to fine-tune the model parameters via Reinforcement Learning (RL).
While effective \citep{stiennon2020learning,bai2022constitutional}, RLHF's complexity and computational demands hinder its widespread adoption.
Moreover, the RL loop necessitates extra model copies - such as for the reward model, and the anchor model used for KL regularization - which significantly increases its memory usage in comparison to standard fine-tuning.

We compare standard RLHF, where all the parameters of the reward model and policy are fine-tuned, to Parameter-Efficient Reinforcement Learning, which leverages LoRA (Low-Rank Adaptation) \citep{hu2021lora} for fine-tuning both the reward model, and the reinforcement learning policy.
While more powerful Parameter Efficient Fine-Tuning (PEFT) and Representation Fine-Tuning (ReFT) approaches have been developed since LoRA, our study focuses on this method, as it is widely adopted.
We hope our results will motivate the benchmarking of other PEFT and ReFT approaches on RLHF tasks.
Despite training only a small fraction of the parameters, we demonstrate that the results obtained with PE-RLHF are on par with those obtained with standard RLHF.

Figure \ref{fig:perl_rl} illustrates the differences between PE-RLHF and standard RLHF.

Our contributions are threefold:
\begin{itemize}
    \item \textbf{Thorough Comparative Analysis:} We conducted an extensive evaluation of PE-RLHF against standard RLHF methods across six diverse datasets and five distinct tasks. While LoRA's efficiency was expected, its surprisingly strong performance establishes it as a superior alternative to full fine-tuning for RLHF. Detailed results are presented in Table \ref{table:all_results}.
    \item \textbf{In-depth Ablation studies:} We systematically examined the influence of LoRA on both RM and RL policy training, considering variations in model size and LoRA ranks.
    \item \textbf{Demonstrated Resource Savings:} We provide empirical measurements demonstrating reductions in memory consumption and training time achieved by PE-RLHF as compared with standard RLHF.
\end{itemize}
We hope that this study will pave the way for more efficient and accessible RLHF, promoting wider adoption and facilitating the development of large models that better align with human preferences.

\begin{table*}
  \caption{PE-RLHF RM training can match standard RM training in terms of reward model accuracy while using 43-74\% of the HBM at peak and trains 1.4-1.9$\times$ faster as compared to standard RM training. PE-RLHF can match the standard RLHF while using 73-80\% HBM at peak, and training 1.15-1.3$\times$ faster.
  }
\begin{threeparttable}
  \centering
  \resizebox{0.9\textwidth}{!}{
  \begin{tabular}{l c c c c c c c}
    \toprule
        \toprule
        \multirow{3}{*}{} & \multirow{3}{*}{} & Harmlessness & Helpfulness & \multicolumn{2}{c}{Summarization} & UI Automation & Visual QA \\
        \cmidrule(lr){3-3} \cmidrule(lr){4-4}  \cmidrule(lr){5-6}  \cmidrule(lr){7-7} \cmidrule(lr){8-8}
        && Anthropic & SHP  & Reddit & Messages & UI Automation & VQAv2 \\
        \midrule
        \midrule
        \multirow{2}{*}{RM \tnote{*}}& Full-tuning & 76.56\% & \textbf{83.2\%} & 78.7\% & - & \textbf{93.12\%} & \textbf{-}   \\
        &  PE-RLHF & \textbf{78.71\%} & 82.2\% & \textbf{79.7\%} & - & 91.8\% &  +0.5\%\tnote{$\dagger$} \\
        \midrule
        \multirow{2}{*}{Efficiency}& PE HBM & 43.1\% & 48\% & 50\% & - & 56\% & 74\% \\
        & Speed-up & $1.6\times$ & $1.5\times$ & $1.7\times$ & - & $1.9\times$ & 1.15$\times$ \\
   \midrule
   \midrule
      \multirow{2}{*}{RL \tnote{**}} & Full-tuning  & 96.6\% & \textbf{63.0\%} & \textbf{87\%} & 73.2\% & 81.6\% & \textbf{+5.5\%}\tnote{$\ddag$} \\
       & PE-RLHF & \textbf{98.2\%} & 61.3\% & 86.5\% & \textbf{75.5\%} & \textbf{86.4\%} & +3.9\%\tnote{$\ddag$} \\
   \midrule
       \multirow{2}{*}{Efficiency}& PE HBM  & 80\% & 75\% & 75\% & 80\% & 74\% & 74\% \\
       & Speed-up & $1.3\times$ & $1.15\times$ & $1.15\times$ & $1.05\times$ & $1.20\times$ & $1.24\times$ \\
    \bottomrule
    \bottomrule
  \end{tabular}}
\begin{tablenotes}\footnotesize
\item[*] Accuracy. 
\item[**] Win rate by a Judge model.
\item[$\dagger$] Absolute win rate change compared to Supervised Fine-Tuning baseline.
\item[$\ddag$] Absolute accuracy change compared to Fully-Tuned baseline.
\end{tablenotes}
\end{threeparttable}
\end{table*}
\label{table:all_results}

\section{Parameter Efficient Reinforcement Learning from Human Feedback}
RLHF involves two phases: reward model training, and reinforcement learning of a policy model.
PE-RLHF applies parameter-efficient fine-tuning techniques to optimize both of these training phases, thus significantly reducing the memory requirements, and increasing the training speed.
We provide a brief overview on RLHF in Appendix \ref{app:rlhf_overview}.

\subsection{Reward Model Training}
\begin{figure}[t]
    \centering
        \includegraphics[width=0.5\textwidth]{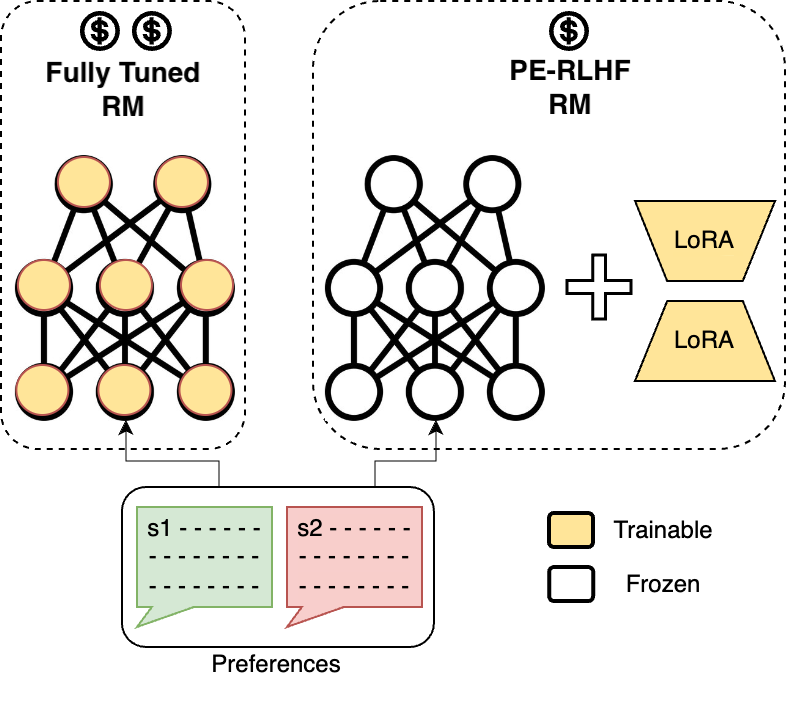}

    \caption{Standard RM training (left) vs. PE-RLHF RM training (right). PE-RLHF RM only trains the LoRA adapters, while keeping the Language Model backbone frozen.}
    \label{fig:perl_rm}
\end{figure}

PE-RLHF constructs reward models as language models with Low-Rank Adaptation (LoRA) adapters.
These adapters are attached to each attention projection matrix within the model.
During training, only the adapters are trained, while the language model backbone remains frozen.
This approach, illustrated in Figure \ref{fig:perl_rm}, significantly reduces the number of trainable parameters.
During inference, the trained LoRA adapters are combined with the projection matrices through a one-time addition operation.
This results in a reward model functionally equivalent to a non-LoRA model, but trained efficiently.

\subsection{Reinforcement Learning of Policy}
\begin{figure}[t]
    \centering
    \includegraphics[width=1.02\linewidth]{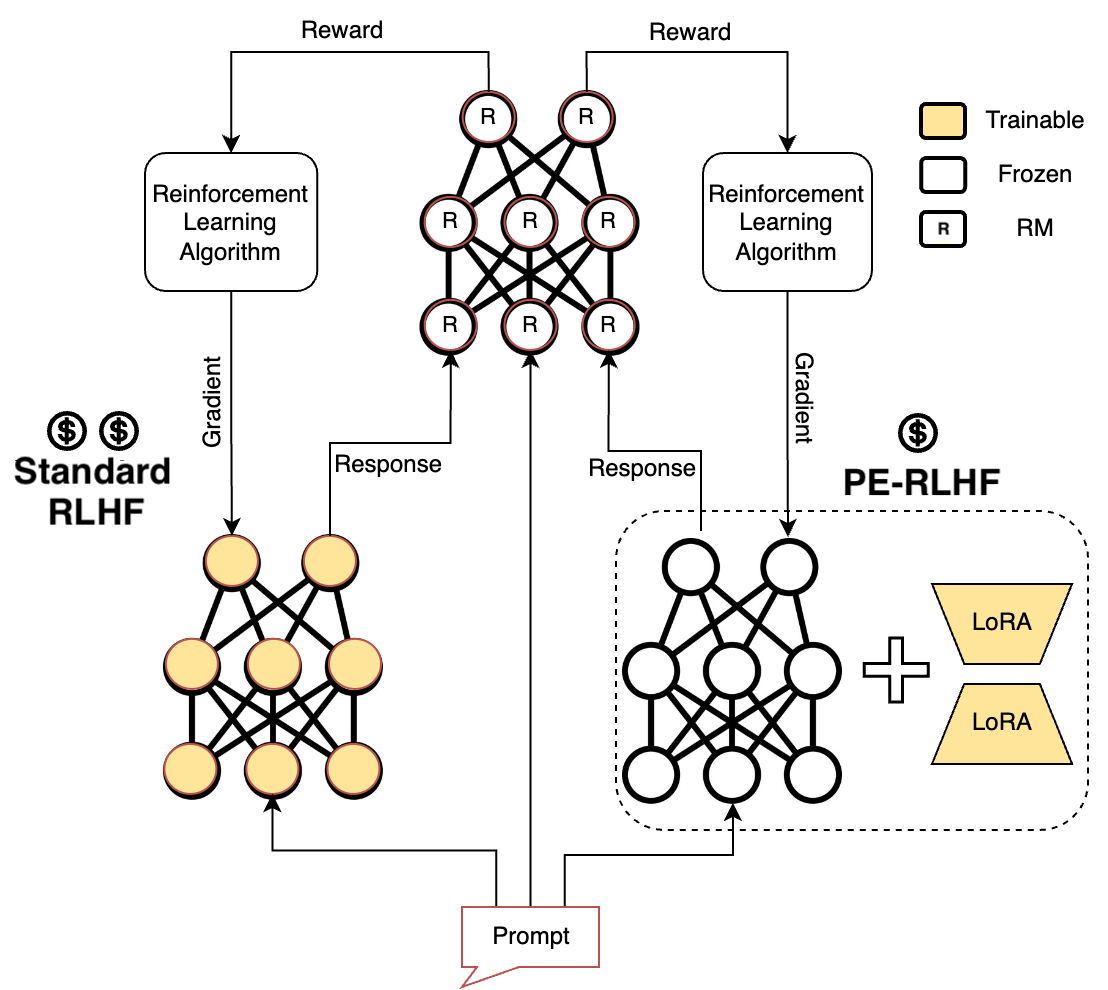}
    \caption{Standard RLHF (left) vs. PE-RLHF right. PE-RLHF only trains the LoRA adapters while keeping the Language Model backbone frozen. 
    }
    \label{fig:perl_rl}
\end{figure}

Similarly, PE-RLHF uses LoRA adapters for policy and value models within the reinforcement learning loop.
As with the reward model, adapters are attached to each attention projection matrix, and trained while keeping the language model backbone frozen (Figure \ref{fig:perl_rl}). 
The policy is then optimized using the policy gradient calculated based on the value model. The value model is trained using the reward score, along with KL regularization with the anchor policy.
We optimize the policy using ``REINFORCE for Language Models'', as used by \citet{lee2023rlaif}.

\section{Datasets and Tasks}

\label{sec:datasets_and_tasks}
We describe the datasets we used for training the reward models, and performing reinforcement learning, categorized by their respective tasks.
Our experiments are based on a diverse collection of datasets, and test various capabilities of language models.
This diversity allows us to evaluate the efficacy PE-RLHF across different domains, and assess its ability to generalize to new tasks.
The datasets are grouped into the following categories:

\paragraph{Text Summarization:} These datasets test a model's ability to condense information, and generate concise summaries.
We experiment with the Reddit TL;DR \citep{volske2017tl,stiennon2020learning}, and BOLT Message Summarization \citep{bolt2018} datasets.
The Reddit TL;DR dataset contains Reddit posts together with human-annotated summaries, used to train models for summarizing these posts.
We filtered this dataset following the work of \citet{stiennon2020learning}.
The BOLT Message Summarization dataset contains chat conversations, used to train models for summarizing sequences of messages.

\paragraph{Harmless Response Generation:} Generating harmless responses is critical to the development of responsible AI systems.
We use the ``Harmlessness" dataset from Anthropic-HH \citep{bai2022training}.
This dataset evaluates a model's ability to generate safe responses to red-teaming prompts.

\paragraph{Helpful Response Generation:} This task asks the model to provide helpful answer for a given question. We use Stanford Human Preference dataset \citep{ethayarajh2022understanding} to evaluate the helpfulness of the generated answer.

\paragraph{UI Automation:} This task assesses the model's ability to understand and interact with user interfaces.
We run our experiments on the AndroidControl dataset \citep{li2024effects}, which consists of human demonstrations of controlling a device, used to train reward models that evaluate the effectiveness of actions in a UI automation task.

\paragraph{Visual Question Answering:} We test a model's ability to understand visual information, and answer questions about it using the VQAv2 dataset \citep{antol2015vqa,goyal2017making}.\\

Detailed information about each dataset can be found in Appendix \ref{app:dataset_details}.

\section{Experimental Setup and Metrics}

\begin{figure*}[t]
    \centering
        \includegraphics[width=0.9\textwidth]{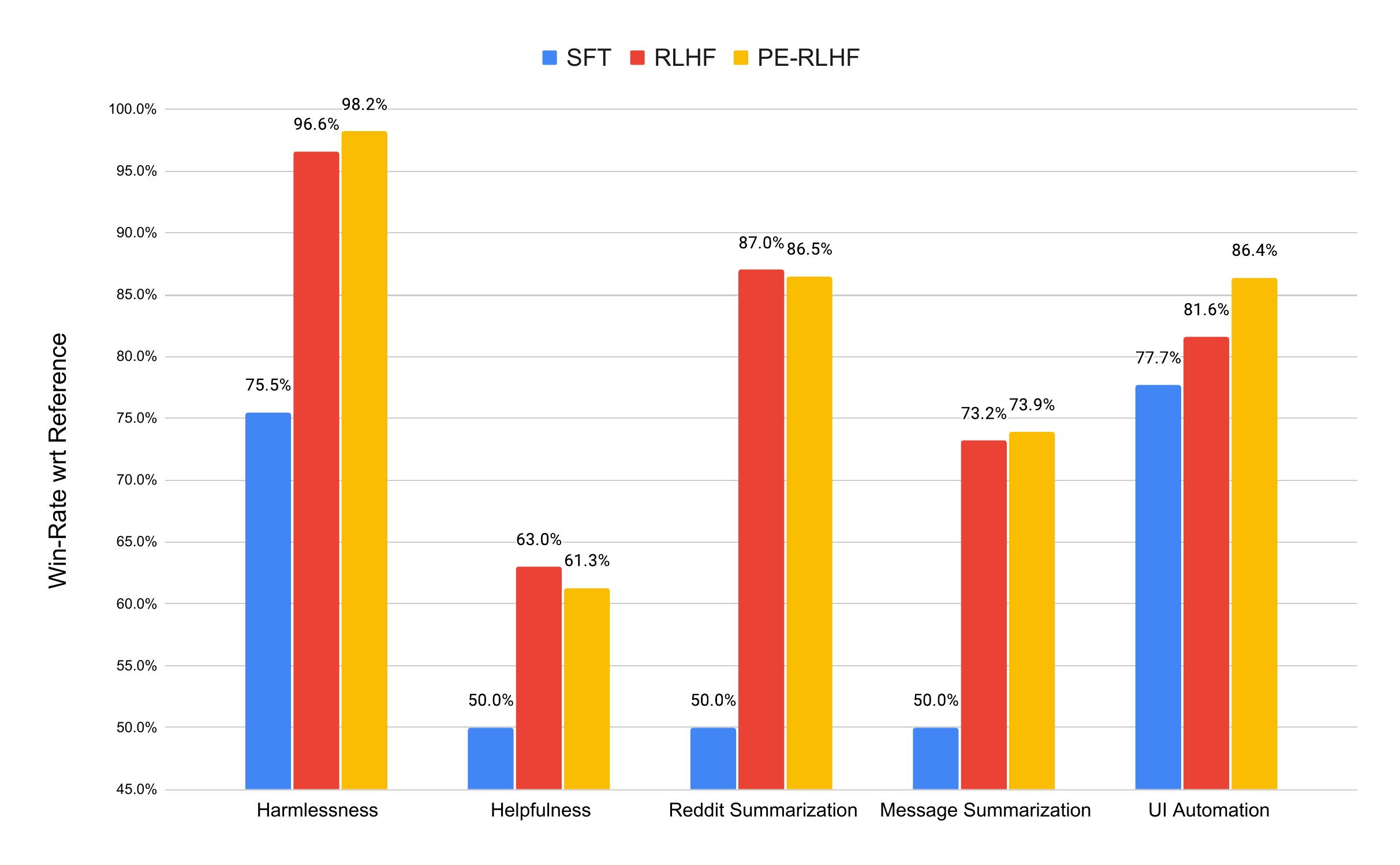}

    \caption{PE-RLHF performs on par with standard RLHF. Both PE-RLHF and RLHF outperform SFT policies significantly on all tasks.}
    \label{fig:perl_chart}
\end{figure*}

We run extensive experiments on six datasets, using models from two different families:
\begin{itemize}
    \item PaLM 2 \citep{palm2techreport}: A text focused large model pretrained following the UL2 \citep{tay2022ul2} paradigm. We experiment with three sizes for this model, referred to as XXS, XS, and S. We use a larger PaLM 2 L as a judge in our evaluations. PaLM 2 models are available via Google Cloud’s Vertex API by the names Gecko (XXS), Otter (XS), Bison (S), and Unicorn (L).
    \item Gemini Pro \citep{geminiteam2023gemini}: A vision language model, also available via Google Cloud’s Vertex API.
\end{itemize}

We emphasize that our experimental setup is independent of the specific models used.

We describe the experiments for reward modeling and reinforcement learning in more details below.

\subsection{Reward Modeling}
We train reward models with the loss described in Appendix \ref{app:rlhf_overview}, and the hyperparameters described in Appendix \ref{app:exp_hp}, varying the configurations of model size, and LoRA rank (which controls the number of trainable parameters).

We evaluate the performance of the preference based reward models using the \textit{pairwise-accuracy}, which measures the proportion of preferred responses ranked higher by the model among pairs of candidate responses.
We evaluate the classification style reward models using the \textit{accuracy}, which indicates whether the reward model score is close to the label of 0 or 1.
We compare the High Bandwidth Memory (HBM) usage as estimated by Jax JIT at the time of training \citep{bradbury2018jax}, and report its peak value.
We also evaluate and report the speed of training for each setting.

\subsection{Reinforcement Learning}
We train reinforcement learning policies using the ``REINFORCE for Language Models" algorithm used by \citet{lee2023rlaif}, using a fixed reward model for each dataset for a fair comparison across the different settings (this is to reduce confounding factors that affect the policy performance).
For every setting we try for policy model, both in size and LoRA rank, we replicate that for the value model as well.
We report the experiment hyperparameters in Appendix \ref{app:exp_hp}.

\subsection{Evaluations}
We evaluate the performance of the RL tuned policies using the PaLM 2 L model (judge model), which is prompted to judge the responses produced by the policy.

We use it to evaluate the quality of the models as follows:
\paragraph{Text Summarization:} We prompt the judge model to pick a preferred response among a pair of baseline, and policy responses.
We report the ``win rate'' of the policy, as defined by the percentage of generated responses that are better than the baseline.
We use an instruction-tuned version of PaLM 2 S to generate baseline responses.
We prompt the judge model twice, with two possible orders, to eliminate positional bias, and average the judgements:
only if an output is preferred in both possible orderings is it considered a win.
If it is only preferred in one ordering, then we label a tie.

\paragraph{Harmless Response Generation:} We prompt the judge LLM to assess whether the generated response is harmless in a YES/NO manner. We report the ``harmless rate'', which measures the fraction of the generated responses that are \textit{harmless}.

\paragraph{Helpful Response Generation:} Similarly to the text summarization task, we prompt the judge model to pick a preferred (more helpful) response among a pair of baseline and policy responses.
We report the ``win rate'' of the policy, as defined by the percentage of generated responses that are better than the baseline. 

\paragraph{Visual Question-Answering:} We evaluate the quality of an RL policy as the difference between its accuracy, and that of a baseline Supervised Fine-Tuned Gemini policy.

\paragraph{UI Automation:} We prompt the judge LLM with the task description, together with the action generated by the policy model. We asked the judge LLM to determine whether the action is correct for the UI automation task.
We calculate the accuracy rate as defined by the percentage of the policy model generated action that are considered as correct ones. \\

We compare and report the speed of each RL training experiment.
In contrast to the speed measured for the reward models, which only include the learning step, the reinforcement learning measurements include episode sampling, reward model scoring, anchor logit calculation, and learning step.

\section{Results and Takeaways}
\subsection{Performance of PE-RLHF}
We observe that PE-RLHF achieves results on par with those of standard RLHF in both reward modeling, and reinforcement learning. 

PE-RLHF RMs match the performance of their RLHF counterparts across diverse tasks, as measured by the accuracy of the reward models on an evaluation split.
These LoRA adapted RMs achieve this performance by training less than $0.1\%$ of the large model's total parameter count.
We summarize the reward modeling results in the top half of Table \ref{table:all_results}.

When guided by the same reward model, PE-RLHF policies achieve performance competitive with those of the policies trained in standard RLHF.
They both perform significantly better than an SFT policy, as seen in Figure \ref{fig:perl_chart}, showing the effectiveness of reinforcement learning.
PE-RLHF policies achieve this performance by training less than $0.1\%$ of the large model's total parameter count for text based tasks, and less than $0.2\%$ of the large model's total parameter count for tasks involving both vision and text.
We summarize the reinforcement learning policy results in the bottom half of Table \ref{table:all_results}.

We assess the quality of PaLM 2 L as a judge by collecting human feedback on 50 samples.
We calculate the agreement between the human and judge model's feedback across tasks, to verify the validity of the latter.
We report details of our verification in Appendix \ref{app:human_corr}.

\begin{table*}[ht]
    \centering
    \caption{PE-RLHF RMs match the accuracy of full-tuned RMs, and get more effective at performing equally to fully-tuned RM as the backbone model increases in size. We don't observe significant changes in performance with varying LoRA ranks.}
    \resizebox{0.85\textwidth}{!}{
    \begin{tabular}{c c  c c  c c c}
        \toprule
        \multirow{3}{*}{Model} & \multirow{3}{*}{Setting} & Harmlessness & Helpfulness & Summarization & UI Automation  \\
        \cmidrule(lr){3-3} \cmidrule(lr){4-4} \cmidrule(lr){5-5}  \cmidrule(lr){6-6}
        && Anthropic & SHP & Reddit & UI Automation \\
        \midrule
        \multirow{5}{*}{PaLM 2 S}
        & Fully-Tuned & 76.6\% & 83.2\% & 78.7\% & 93.1\%  \\
        & LoRA 16 & 75.0\% & 81.3\% & 77.0\%     & 92.2\% \\
        & LoRA 8 & 79.1\% & 81.6\% & 77.3\%      & 92.0\% \\
        & LoRA 4 & 78.7\% & 82.2\% & 79.7\%      & 91.2\% \\
        & LoRA 1 & 76.0\% & 80.9\% & 77.2\%      & 90.8\% \\
        \midrule 
        \multirow{2}{*}{PaLM 2 XS}
        &Fully-Tuned & 77.0\% & 82.0\% & 78.1\%   & 88.6\% \\
        &LoRA 8 & 76.8\% & 82.6\% & -            & 89.2\%\\
        &LoRA 4 & - & 81.8\% & 76.8\%            &  89.7\% \\
        \midrule 
        \multirow{2}{*}{PaLM 2 XXS}
        &Fully-Tuned & 73.4\% & 80.2\% & 75.4\% & 84.4\% \\
        &LoRA 8 & 72.5\% & 79.3\% & - & 87.1\% \\
        &LoRA 4 & - & 77.8\% & 73.2\% & 85.8\% \\
        \bottomrule
    \end{tabular}
    }
    \\
    \vspace{0.6em}
    \resizebox{0.57\textwidth}{!}{
    \begin{threeparttable}
    \begin{tabular}{c c  c}
        \toprule
        \multirow{2}{*}{Model} &  \multirow{2}{*}{Setting} & Visual Question Answering \\
        \cmidrule(lr){3-3} 
        && VQAv2\tnote{*} \\
        \midrule
        \multirow{6}{*}{Gemini Pro}
        &Fully-Tuned & 0.0\% \\
        &LoRA 32 & -1.2\% \\
        &LoRA 16 & -1.0\% \\
        &LoRA 8 &  -1.1\% \\
        &LoRA 4 &  -0.6\% \\
        &LoRA 1 &  -1.0\% \\
        \bottomrule
    \end{tabular}
    \begin{tablenotes}\footnotesize
    \item[*] Absolute accuracy change compared to Fully-Tuned baseline.
    \end{tablenotes}
    \end{threeparttable}
    }
    \label{table:rm_results}
\end{table*}

\subsection{Effects of Model Size and LoRA Rank}
We conduct ablation studies across model size and LoRA rank to test their effect on the performance of the reward model and the reinforcement learning policy. 

We observe that changing the LoRA rank does not significantly affect the performance of the reward models.
However, PE-RLHF is more effective at modeling reward, and performs closer to standard full-tuning when the size of the model backbone increases.
At the lowest model size, we see PE-RLHF falling marginally short of the fully-tuned reward models, whereas it matches the performance of fully-tuned ones for the largest model sizes.

In the reinforcement learning experiments, on the other hand, we observe that the PE-RLHF policies perform better (as evaluated by the judge LLM) as the LoRA rank increases.
Bigger model backbones translate into better results for the PE-RLHF policies in comparison to the standard RLHF ones, consistent with the trend observed for the reward models.
PE-RLHF falls marginally short of the fully-tuned RL policies with the smallest model size, but matches the performance of RL policies with the biggest size.
See Tables \ref{table:rm_results} and \ref{table:rl_results} for detailed results.

\begin{table*}[h]
    \centering
    \caption{PE-RLHF policies match the performance of standard RL policies, and become more effective as the LM increases in size. We don't observe significant variations in performance with the LoRA rank.}
    \resizebox{0.85\textwidth}{!}{
    \begin{tabular}{c c  c c c c c }
        \toprule
        \multirow{3}{*}{Model} & \multirow{3}{*}{Setting} & Harmlessness & Helpfulness & \multicolumn{2}{c}{Summarization} & UI Automation  \\
        \cmidrule(lr){3-3} \cmidrule(lr){4-4} \cmidrule(lr){5-6} \cmidrule(lr){7-7}
        && Anthropic & SHP & Reddit & Messages & AndroidControl \\
        \midrule
        \multirow{5}{*}{PaLM 2 S}
        & Fully-Tuned & 96.6\% & 63.0\% & 87\% & 73.2\% & 81.6\%  \\
        & LoRA 16 & 96.4\% & 61.3\% & 86.5\% & 75.5\% & 86.4\% \\
        & LoRA 8 & 97.7\% & 58.3\% & 85.7\% & 73.4\%  & 85.4\% \\
        & LoRA 4 & 96.7\% & 60.1\% & 84.2\% & 73.9\%  & 84.5\% \\
        & LoRA 1 & 98.2\% & 57.9\% & 85\% & 73.1\%  & 77.7\%\\
        & SFT & 75.5\% & 50\% & 50\% & 50\%  & 77.7\% \\
        \midrule 
        \multirow{2}{*}{PaLM 2 XS}
        &Fully-Tuned & 96.1\% & 53.5\% & 77.7\% & 64.3\% & 52.4\%\\
        &LoRA 16 & 97.4\% & 52.5\% & 79.5\% & 65.5\%  & 69.4\%\\
        & SFT & 70.3\% & - & 33.6\% & 48.1\% & 48.5\% \\
        \midrule 
        \multirow{2}{*}{PaLM 2 XXS}
        & Fully-Tuned & 96.6\% & 5.92\% & 48.4\% & 33.7\%  & 11.7\% \\
        & LoRA 16 & 96.6\% & 5.61\% & 31.7\% & 31.7\% & 14.6\% \\
        & SFT & 66.7\% & - & 16.0\% & 14.1\% & 24.3\% \\
        \bottomrule
    \end{tabular}
    }
    \\
    \vspace{0.6em}
    \resizebox{0.57\textwidth}{!}{
    \begin{threeparttable}
    \begin{tabular}{c c  c}
        \toprule
        \multirow{2}{*}{Model} &  \multirow{2}{*}{Setting} & Visual Question Answering \\
        \cmidrule(lr){3-3} 
        && VQAv2\tnote{*} \\
        \midrule
        \multirow{7}{*}{Gemini Pro}
        &Fully-Tuned & +5.5\% \\
        &LoRA 32 &  +3.9\% \\
        &LoRA 16 &  +3.1\% \\
        &LoRA 8 &  +2.5\% \\
        &LoRA 4 &  +3.0\% \\
        &LoRA 1 & +3.0\% \\
        &SFT & +0.0\% \\
        \bottomrule
    \end{tabular}
    \begin{tablenotes}\footnotesize
    \item[*] Absolute accuracy change compared to SFT.
    \end{tablenotes}
    \end{threeparttable}
    }
    \label{table:rl_results}
\end{table*}

\subsection{Memory and Speed Advantages of PE-RLHF}
Modern optimizers, such as Adam \citep{kingma2017adam} and Adafactor \citep{shazeer2018adafactor}, require substantial memory to track various factors for each trainable parameter.
By largely reducing the number of trainable parameters, PE-RLHF significantly lowers the memory footprint of both reward model training and reinforcement learning of policy.
PE-RLHF RM training achieves the performance of a fully-tuned RM, while using only 43\% to 74\% of the peak HBM it needs for training.
PE-RLHF reinforcement learning achieves the performance of standard RLHF, while using only 74\% to 80\% of the peak HBM.

The reduction in number of trainable parameters also translates into significant training speed-ups, as fewer parameters need updating at each training step.
PE-RLHF RM trains at 1.15 $\times$ to 1.9$\times$ the training speed of standard RM.
PE-RLHF RL trains 1.15$\times$ to 1.24$\times$ the speed of the RL loop in standard RLHF. 
We note that the LoRA reward models and policies converge in a similar number of steps as the fully tuned ones, so that the speed-ups in training steps translate into faster runs.

We observe that the memory savings, and training speed-up do not vary significantly with the LoRA rank, since the change in trainable parameters is extremely small in comparison to total parameters (<1\% in maximum LoRA rank of 32).
We report exact numbers for memory savings, and training speed-up in Table \ref{table:all_results}.
We also note that the memory savings and speed-up depend on multiple factors, such as sequence lengths of the examples, the accelerators being used, etc.

\section{Related Work}
\subsection{Pretrained Large Models (PLMs)}
PLMs like LaMDA \citep{thoppilan2022lamda}, BLOOM \citep{workshop2022bloom}, PaLM \citep{chowdhery2022palm, palm2techreport}, and GPT-4 \citep{openai2023gpt4} have demonstrated remarkable performance across diverse tasks, including summarization \citep{stiennon2020learning}, instruction following \citep{ouyang2022training,lai2023okapi}, and dialogue generation \citep{friedman2023leveraging, jandaghi2023faithful}.
Despite their success, limitations such as factual inaccuracies and imperfect instruction adherence remain \citep{radford2018improving, ramachandran2016unsupervised, wei2021finetuned}.

\subsection{Aligning PLMs with Human/AI Preferences}
To address these limitations, aligning PLMs with human preferences has emerged as a crucial research area \citep{christiano2017deep, leike2018scalable, wang2023aligning, ji2023ai}.
This typically involves collecting preference data on generated outputs, and fine-tuning the model with reward functions based on this data.
However, overfitting the reward function can be problematic \citep{azar2023general}, requiring techniques like early stopping or parameter reduction to ensure optimal policy training.
While most alignment research focuses on natural language tasks, recent efforts explore its application in other modalities like vision and audio \citep{lee2023aligning, sun2023aligning}.

\subsection{Techniques for Alignment}
Several techniques have been developed for aligning PLMs, such as Reward rAnked Fine-Tuning (RAFT) \citep{dong2023raft}, RRHF \citep{yuan2023rrhf}, Reinforcement Learning from Human Feedback (RLHF) \citep{christiano2017deep, ouyang2022training, azar2023general}, Direct Preference Optimization (DPO) \citep{rafailov2023direct, azar2023general}, Sequence Likelihood Calibration with Human Feedback (SLIC-HF) \citep{zhao2023slichf}, Pairwise Cringe Optimization \citep{xu2023things}, and Self-Rewarding Language Models \citep{yuan2024selfrewarding}.
Among these, RLHF is particularly popular.
This work explores combining RLHF with parameter-efficient methods to improve computational and memory efficiency.

\subsection{Parameter-Efficient Fine-Tuning (PEFT), Representation Fine-Tuning (ReFT), and LoRA}
Parameter-Efficient Fine-Tuning (PEFT) methods, like LoRA \citep{hu2021lora} and DoRA \citep{liu2024dora}, and Representation Fine-Tuning (ReFT) ones \cite{wu2024reft} reduce the number of trainable parameters in PLMs, while maintaining comparable performance to full fine-tuning.
These methods are crucial for adapting PLMs to downstream tasks with limited data and computational resources.
LoRA specifically factorizes weight updates into low-rank matrices, significantly reducing the number of parameters to be trained.
To the best of our knowledge, there hasn't been any work prior to ours extensively benchmarking parameter efficient approaches for RLHF.

\subsection{Infrastructure and Implementation}
While the Transformer Reinforcement Learning (TRL) library \citep{vonwerra2022trl} offers similar functionalities, its multi-adapter RL feature remains experimental and lacks support for parallelization and vision modalities.
This work is implemented using the PAX \citep{paxml} and SeqIO \citep{roberts2022scaling} libraries, along with a custom-designed RL training loop infrastructure.

\section{Conclusion and Future Work}
\label{sec:conclusion}
This work demonstrates, through extensive experiments, that the Parameter-Efficient Reinforcement Learning from Human Feedback (PE-RLHF) setup that leverages LoRA achieves comparable performance to standard RLHF, while significantly reducing memory usage and training time.
Specifically, PE-RLHF reduces peak memory usage by approximately 50\%, and induces a speed up of up to 90\% in the reward model training.
While the RL loop shows more modest gains of 27\% peak memory reduction, and a 30\% speed-up, these improvements still contribute to a more efficient training process.

While PE-RLHF demonstrates success in matching the performance of standard RLHF on in-domain test sets, further investigation is needed to explore its generalizability.
Since parameter efficient fine-tuning methods can be prone to over-fitting and ``reward-hacking", we propose three avenues for future work:
\begin{itemize}
    \item \textbf{Broader Generalization}: Ensemble models like Mixture-of-LoRA \citep{wu2024mixture} could enhance cross-domain generalization by introducing robustness during training. This approach holds promise for achieving broader applicability without significant computational overhead.
    \item \textbf{Mitigating Reward Hacking}: Reward models are susceptible to ``reward hacking", where the model exploits loopholes in the reward function instead of learning the desired behavior. Recent research \citep{rame2024warm} suggests that weight-averaging models can mitigate this issue. Integrating such techniques with PE-RLHF's efficient adapter-based approach could offer similar benefits at a lower computational cost.
    \item \textbf{Open Sourcing}: We aim to share examples comparing PE-RLHF to standard RLHF using open-source models.
\end{itemize}

By addressing these avenues in future work, PE-RLHF has the potential to become a powerful and efficient tool for training large language and vision-language models with reinforcement learning, paving the way for broader and robust applications.

\section*{Limitations}
While the PE-RLHF setup demonstrates promising results in terms of efficiency and performance, there are certain limitations that warrant further investigation

\begin{itemize}
    \item \textbf{Potential Overfitting:} As with any parameter-efficient fine-tuning method, there is a risk of overfitting to the training data. We describe ideas in Section \ref{sec:conclusion} to explore regularization techniques and other mitigation strategies for this potential issue.
    
    \item \textbf{Data Efficiency:} Although PE-RLHF reduces the computational burden of RLHF, we are still not sure about the optimal data requirement of the RLHF process. Exploring data-efficient approaches for PE-RLHF is an important direction for future research.
    
    \item \textbf{Comparison with Other PEFT and ReFT Methods:} We solely focus on LoRA as the parameter-efficient fine-tuning (PEFT) method. While we expect other PEFT methods, like DoRA, and ReFT ones to behave similarly, our benchmarking work does not include these newer methods, and there is room to explore their behaviour in the RLHF setup, and whether their improvements over LoRA translate into better alignment with human preferences.
\end{itemize}

Despite these limitations, we believe that our work contributes to the development of more efficient and accessible alignment techniques for large language and vision-language models. Researching along the directions mentioned above will further strengthen the applicability and effectiveness of PE-RLHF across broader scenarios.

\section*{Ethics Statement}
While reducing alignment barriers for LLMs holds immense promise, it also presents ethical challenges by potentially facilitating malicious applications.
The PE-RLHF method, while enabling efficient alignment, could be misused to train LLMs for generating harmful content like misinformation or hate speech at a reduced cost.
Mitigating this risk necessitates careful governance of powerful LLMs, with controlled use to restrict "white-box" access to prevent malicious actors from manipulating the model's inner workings, robust monitoring systems to detect and prevent the generation of harmful content, and the development and enforcement of ethical guidelines for LLM development and deployment.
\\

\section*{Reproducibility}
To ensure the reproducibility of our findings, we provide comprehensive details throughout the paper and Appendix. Specifically:
\begin{itemize}
    \item \textbf{Datasets:} We describe all open-sourced datasets in Appendix \ref{app:dataset_details}.
    \item \textbf{Model Training and Experiments:} Appendix \ref{app:exp_hp} details the model training procedure and experimental setup.
    \item \textbf{Evaluation:} We outline the automated evaluation methodology in Appendix \ref{app:human_corr}, and present the evaluation prompts in Table \ref{tab:judge_prompts}.
    \item \textbf{Benchmarking Models:} Our benchmarking experiments are conducted using PaLM 2 and Gemini models, accessible via Google Cloud's Vertex API. We provide details to facilitate replication using other open-sourced models.
\end{itemize}
This documentation enables researchers to reproduce our experiments and verify our results using alternative models and configurations.

\section*{Acknowledgements}
\label{sec:acknowledgements}
We thank the individuals who designed and built the RL training infrastructure used in this paper: Léonard Hussenot, Johan Ferret, Robert Dadashi, Geoffrey Cideron, Alexis Jacq, Sabela Ramos, Piotr Stanczyk, Sertan Girgin, Danila Sinopalnikov, Amélie Héliou, Nikola Momchev, and Olivier Bachem.
Léonard Hussenot was instrumental in helping us build our first parameter efficient RL training pipelines.

We thank Ahmad Beirami for useful conversations that led to the idea of using PERL as an efficient ensembling method as future work, and Yun Zhu for useful discussions about some of the datasets used.

Finally, the authors would like to thank Alex Castro-Ros, Kevin Robinson, and Kathy Meier-Hellstern for their careful reviews and many suggestions to improve our initial manuscript.

\bibliography{emnlp2024_conference}

\begin{thebibliography}{62}
\providecommand{\natexlab}[1]{#1}

\bibitem[{Amodei et~al.(2016)Amodei, Olah, Steinhardt, Christiano, Schulman,
  and Man{\'e}}]{amodei2016concrete}
Dario Amodei, Chris Olah, Jacob Steinhardt, Paul Christiano, John Schulman, and
  Dan Man{\'e}. 2016.
\newblock Concrete problems in ai safety.
\newblock \emph{arXiv preprint arXiv:1606.06565}.

\bibitem[{Anil et~al.(2023)Anil, Dai, Firat, Johnson, Lepikhin, Passos,
  Shakeri, Taropa, Bailey, Chen, Chu, Clark, Shafey, Huang, Meier-Hellstern,
  Mishra, Moreira, Omernick, Robinson, Ruder, Tay, Xiao, Xu, Zhang, Abrego,
  Ahn, Austin, Barham, Botha, Bradbury, Brahma, Brooks, Catasta, Cheng, Cherry,
  Choquette-Choo, Chowdhery, Crepy, Dave, Dehghani, Dev, Devlin, Díaz, Du,
  Dyer, Feinberg, Feng, Fienber, Freitag, Garcia, Gehrmann, Gonzalez, Gur-Ari,
  Hand, Hashemi, Hou, Howland, Hu, Hui, Hurwitz, Isard, Ittycheriah, Jagielski,
  Jia, Kenealy, Krikun, Kudugunta, Lan, Lee, Lee, Li, Li, Li, Li, Li, Lim, Lin,
  Liu, Liu, Maggioni, Mahendru, Maynez, Misra, Moussalem, Nado, Nham, Ni,
  Nystrom, Parrish, Pellat, Polacek, Polozov, Pope, Qiao, Reif, Richter, Riley,
  Ros, Roy, Saeta, Samuel, Shelby, Slone, Smilkov, So, Sohn, Tokumine, Valter,
  Vasudevan, Vodrahalli, Wang, Wang, Wang, Wang, Wieting, Wu, Xu, Xu, Xue, Yin,
  Yu, Zhang, Zheng, Zheng, Zhou, Zhou, Petrov, and Wu}]{palm2techreport}
Rohan Anil, Andrew~M. Dai, Orhan Firat, Melvin Johnson, Dmitry Lepikhin,
  Alexandre Passos, Siamak Shakeri, Emanuel Taropa, Paige Bailey, Zhifeng Chen,
  Eric Chu, Jonathan~H. Clark, Laurent~El Shafey, Yanping Huang, Kathy
  Meier-Hellstern, Gaurav Mishra, Erica Moreira, Mark Omernick, Kevin Robinson,
  Sebastian Ruder, Yi~Tay, Kefan Xiao, Yuanzhong Xu, Yujing Zhang,
  Gustavo~Hernandez Abrego, Junwhan Ahn, Jacob Austin, Paul Barham, Jan Botha,
  James Bradbury, Siddhartha Brahma, Kevin Brooks, Michele Catasta, Yong Cheng,
  Colin Cherry, Christopher~A. Choquette-Choo, Aakanksha Chowdhery, Clément
  Crepy, Shachi Dave, Mostafa Dehghani, Sunipa Dev, Jacob Devlin, Mark Díaz,
  Nan Du, Ethan Dyer, Vlad Feinberg, Fangxiaoyu Feng, Vlad Fienber, Markus
  Freitag, Xavier Garcia, Sebastian Gehrmann, Lucas Gonzalez, Guy Gur-Ari,
  Steven Hand, Hadi Hashemi, Le~Hou, Joshua Howland, Andrea Hu, Jeffrey Hui,
  Jeremy Hurwitz, Michael Isard, Abe Ittycheriah, Matthew Jagielski, Wenhao
  Jia, Kathleen Kenealy, Maxim Krikun, Sneha Kudugunta, Chang Lan, Katherine
  Lee, Benjamin Lee, Eric Li, Music Li, Wei Li, YaGuang Li, Jian Li, Hyeontaek
  Lim, Hanzhao Lin, Zhongtao Liu, Frederick Liu, Marcello Maggioni, Aroma
  Mahendru, Joshua Maynez, Vedant Misra, Maysam Moussalem, Zachary Nado, John
  Nham, Eric Ni, Andrew Nystrom, Alicia Parrish, Marie Pellat, Martin Polacek,
  Alex Polozov, Reiner Pope, Siyuan Qiao, Emily Reif, Bryan Richter, Parker
  Riley, Alex~Castro Ros, Aurko Roy, Brennan Saeta, Rajkumar Samuel, Renee
  Shelby, Ambrose Slone, Daniel Smilkov, David~R. So, Daniel Sohn, Simon
  Tokumine, Dasha Valter, Vijay Vasudevan, Kiran Vodrahalli, Xuezhi Wang,
  Pidong Wang, Zirui Wang, Tao Wang, John Wieting, Yuhuai Wu, Kelvin Xu, Yunhan
  Xu, Linting Xue, Pengcheng Yin, Jiahui Yu, Qiao Zhang, Steven Zheng,
  Ce~Zheng, Weikang Zhou, Denny Zhou, Slav Petrov, and Yonghui Wu. 2023.
\newblock \href {https://arxiv.org/abs/2305.10403} {Palm 2 technical report}.
\newblock \emph{Preprint}, arXiv:2305.10403.

\bibitem[{Anthropic(2022)}]{hf2022anthropic}
Anthropic. 2022.
\newblock \emph{The Anthropic-HH Dataset}.
\newblock \url{https://huggingface.co/datasets/Anthropic/hh-rlhf} [Accessed:
  2024-01-03].

\bibitem[{Antol et~al.(2015)Antol, Agrawal, Lu, Mitchell, Batra, Zitnick, and
  Parikh}]{antol2015vqa}
Stanislaw Antol, Aishwarya Agrawal, Jiasen Lu, Margaret Mitchell, Dhruv Batra,
  C~Lawrence Zitnick, and Devi Parikh. 2015.
\newblock Vqa: Visual question answering.
\newblock In \emph{Proceedings of the IEEE international conference on computer
  vision}, pages 2425--2433.

\bibitem[{Azar et~al.(2023)Azar, Rowland, Piot, Guo, Calandriello, Valko, and
  Munos}]{azar2023general}
Mohammad~Gheshlaghi Azar, Mark Rowland, Bilal Piot, Daniel Guo, Daniele
  Calandriello, Michal Valko, and Rémi Munos. 2023.
\newblock \href {https://arxiv.org/abs/2310.12036} {A general theoretical
  paradigm to understand learning from human preferences}.
\newblock \emph{Preprint}, arXiv:2310.12036.

\bibitem[{Bai et~al.(2022{\natexlab{a}})Bai, Jones, Ndousse, Askell, Chen,
  DasSarma, Drain, Fort, Ganguli, Henighan, Joseph, Kadavath, Kernion, Conerly,
  El-Showk, Elhage, Hatfield-Dodds, Hernandez, Hume, Johnston, Kravec, Lovitt,
  Nanda, Olsson, Amodei, Brown, Clark, McCandlish, Olah, Mann, and
  Kaplan}]{bai2022training}
Yuntao Bai, Andy Jones, Kamal Ndousse, Amanda Askell, Anna Chen, Nova DasSarma,
  Dawn Drain, Stanislav Fort, Deep Ganguli, Tom Henighan, Nicholas Joseph,
  Saurav Kadavath, Jackson Kernion, Tom Conerly, Sheer El-Showk, Nelson Elhage,
  Zac Hatfield-Dodds, Danny Hernandez, Tristan Hume, Scott Johnston, Shauna
  Kravec, Liane Lovitt, Neel Nanda, Catherine Olsson, Dario Amodei, Tom Brown,
  Jack Clark, Sam McCandlish, Chris Olah, Ben Mann, and Jared Kaplan.
  2022{\natexlab{a}}.
\newblock \href {https://arxiv.org/abs/2204.05862} {Training a helpful and
  harmless assistant with reinforcement learning from human feedback}.
\newblock \emph{Preprint}, arXiv:2204.05862.

\bibitem[{Bai et~al.(2022{\natexlab{b}})Bai, Kadavath, Kundu, Askell, Kernion,
  Jones, Chen, Goldie, Mirhoseini, McKinnon et~al.}]{bai2022constitutional}
Yuntao Bai, Saurav Kadavath, Sandipan Kundu, Amanda Askell, Jackson Kernion,
  Andy Jones, Anna Chen, Anna Goldie, Azalia Mirhoseini, Cameron McKinnon,
  et~al. 2022{\natexlab{b}}.
\newblock Constitutional ai: Harmlessness from ai feedback.
\newblock \emph{arXiv preprint arXiv:2212.08073}.

\bibitem[{Bommasani et~al.(2022)Bommasani, Hudson, Adeli, Altman, Arora, von
  Arx, Bernstein, Bohg, Bosselut, Brunskill, Brynjolfsson, Buch, Card,
  Castellon, Chatterji, Chen, Creel, Davis, Demszky, Donahue, Doumbouya,
  Durmus, Ermon, Etchemendy, Ethayarajh, Fei-Fei, Finn, Gale, Gillespie, Goel,
  Goodman, Grossman, Guha, Hashimoto, Henderson, Hewitt, Ho, Hong, Hsu, Huang,
  Icard, Jain, Jurafsky, Kalluri, Karamcheti, Keeling, Khani, Khattab, Koh,
  Krass, Krishna, Kuditipudi, Kumar, Ladhak, Lee, Lee, Leskovec, Levent, Li,
  Li, Ma, Malik, Manning, Mirchandani, Mitchell, Munyikwa, Nair, Narayan,
  Narayanan, Newman, Nie, Niebles, Nilforoshan, Nyarko, Ogut, Orr,
  Papadimitriou, Park, Piech, Portelance, Potts, Raghunathan, Reich, Ren, Rong,
  Roohani, Ruiz, Ryan, Ré, Sadigh, Sagawa, Santhanam, Shih, Srinivasan,
  Tamkin, Taori, Thomas, Tramèr, Wang, Wang, Wu, Wu, Wu, Xie, Yasunaga, You,
  Zaharia, Zhang, Zhang, Zhang, Zhang, Zheng, Zhou, and
  Liang}]{bommasani2022opportunities}
Rishi Bommasani, Drew~A. Hudson, Ehsan Adeli, Russ Altman, Simran Arora, Sydney
  von Arx, Michael~S. Bernstein, Jeannette Bohg, Antoine Bosselut, Emma
  Brunskill, Erik Brynjolfsson, Shyamal Buch, Dallas Card, Rodrigo Castellon,
  Niladri Chatterji, Annie Chen, Kathleen Creel, Jared~Quincy Davis, Dora
  Demszky, Chris Donahue, Moussa Doumbouya, Esin Durmus, Stefano Ermon, John
  Etchemendy, Kawin Ethayarajh, Li~Fei-Fei, Chelsea Finn, Trevor Gale, Lauren
  Gillespie, Karan Goel, Noah Goodman, Shelby Grossman, Neel Guha, Tatsunori
  Hashimoto, Peter Henderson, John Hewitt, Daniel~E. Ho, Jenny Hong, Kyle Hsu,
  Jing Huang, Thomas Icard, Saahil Jain, Dan Jurafsky, Pratyusha Kalluri,
  Siddharth Karamcheti, Geoff Keeling, Fereshte Khani, Omar Khattab, Pang~Wei
  Koh, Mark Krass, Ranjay Krishna, Rohith Kuditipudi, Ananya Kumar, Faisal
  Ladhak, Mina Lee, Tony Lee, Jure Leskovec, Isabelle Levent, Xiang~Lisa Li,
  Xuechen Li, Tengyu Ma, Ali Malik, Christopher~D. Manning, Suvir Mirchandani,
  Eric Mitchell, Zanele Munyikwa, Suraj Nair, Avanika Narayan, Deepak
  Narayanan, Ben Newman, Allen Nie, Juan~Carlos Niebles, Hamed Nilforoshan,
  Julian Nyarko, Giray Ogut, Laurel Orr, Isabel Papadimitriou, Joon~Sung Park,
  Chris Piech, Eva Portelance, Christopher Potts, Aditi Raghunathan, Rob Reich,
  Hongyu Ren, Frieda Rong, Yusuf Roohani, Camilo Ruiz, Jack Ryan, Christopher
  Ré, Dorsa Sadigh, Shiori Sagawa, Keshav Santhanam, Andy Shih, Krishnan
  Srinivasan, Alex Tamkin, Rohan Taori, Armin~W. Thomas, Florian Tramèr,
  Rose~E. Wang, William Wang, Bohan Wu, Jiajun Wu, Yuhuai Wu, Sang~Michael Xie,
  Michihiro Yasunaga, Jiaxuan You, Matei Zaharia, Michael Zhang, Tianyi Zhang,
  Xikun Zhang, Yuhui Zhang, Lucia Zheng, Kaitlyn Zhou, and Percy Liang. 2022.
\newblock \href {https://arxiv.org/abs/2108.07258} {On the opportunities and
  risks of foundation models}.
\newblock \emph{Preprint}, arXiv:2108.07258.

\bibitem[{Bradbury et~al.(2018)Bradbury, Frostig, Hawkins, Johnson, Leary,
  Maclaurin, Necula, Paszke, Vander{P}las, Wanderman-{M}ilne, and
  Zhang}]{bradbury2018jax}
James Bradbury, Roy Frostig, Peter Hawkins, Matthew~James Johnson, Chris Leary,
  Dougal Maclaurin, George Necula, Adam Paszke, Jake Vander{P}las, Skye
  Wanderman-{M}ilne, and Qiao Zhang. 2018.
\newblock \href {http://github.com/google/jax} {{JAX}: composable
  transformations of {P}ython+{N}um{P}y programs}.

\bibitem[{Bradley and Terry(1952)}]{bradley-terry}
Ralph~Allan Bradley and Milton~E. Terry. 1952.
\newblock \href {https://api.semanticscholar.org/CorpusID:125209808} {Rank
  analysis of incomplete block designs: I. the method of paired comparisons}.
\newblock \emph{Biometrika}, 39:324.

\bibitem[{Chen et~al.(2018)}]{bolt2018}
Song Chen et~al. 2018.
\newblock Bolt english sms/chat ldc2018t19.
\newblock Web Download.
\newblock Philadelphia: Linguistic Data Consortium.

\bibitem[{Chowdhery et~al.(2022)Chowdhery, Narang, Devlin, Bosma, Mishra,
  Roberts, Barham, Chung, Sutton, Gehrmann, Schuh, Shi, Tsvyashchenko, Maynez,
  Rao, Barnes, Tay, Shazeer, Prabhakaran, Reif, Du, Hutchinson, Pope, Bradbury,
  Austin, Isard, Gur-Ari, Yin, Duke, Levskaya, Ghemawat, Dev, Michalewski,
  Garcia, Misra, Robinson, Fedus, Zhou, Ippolito, Luan, Lim, Zoph, Spiridonov,
  Sepassi, Dohan, Agrawal, Omernick, Dai, Pillai, Pellat, Lewkowycz, Moreira,
  Child, Polozov, Lee, Zhou, Wang, Saeta, Diaz, Firat, Catasta, Wei,
  Meier-Hellstern, Eck, Dean, Petrov, and Fiedel}]{chowdhery2022palm}
Aakanksha Chowdhery, Sharan Narang, Jacob Devlin, Maarten Bosma, Gaurav Mishra,
  Adam Roberts, Paul Barham, Hyung~Won Chung, Charles Sutton, Sebastian
  Gehrmann, Parker Schuh, Kensen Shi, Sasha Tsvyashchenko, Joshua Maynez,
  Abhishek Rao, Parker Barnes, Yi~Tay, Noam Shazeer, Vinodkumar Prabhakaran,
  Emily Reif, Nan Du, Ben Hutchinson, Reiner Pope, James Bradbury, Jacob
  Austin, Michael Isard, Guy Gur-Ari, Pengcheng Yin, Toju Duke, Anselm
  Levskaya, Sanjay Ghemawat, Sunipa Dev, Henryk Michalewski, Xavier Garcia,
  Vedant Misra, Kevin Robinson, Liam Fedus, Denny Zhou, Daphne Ippolito, David
  Luan, Hyeontaek Lim, Barret Zoph, Alexander Spiridonov, Ryan Sepassi, David
  Dohan, Shivani Agrawal, Mark Omernick, Andrew~M. Dai,
  Thanumalayan~Sankaranarayana Pillai, Marie Pellat, Aitor Lewkowycz, Erica
  Moreira, Rewon Child, Oleksandr Polozov, Katherine Lee, Zongwei Zhou, Xuezhi
  Wang, Brennan Saeta, Mark Diaz, Orhan Firat, Michele Catasta, Jason Wei,
  Kathy Meier-Hellstern, Douglas Eck, Jeff Dean, Slav Petrov, and Noah Fiedel.
  2022.
\newblock \href {https://arxiv.org/abs/2204.02311} {Palm: Scaling language
  modeling with pathways}.
\newblock \emph{Preprint}, arXiv:2204.02311.

\bibitem[{Christiano et~al.(2017)Christiano, Leike, Brown, Martic, Legg, and
  Amodei}]{christiano2017deep}
Paul~F Christiano, Jan Leike, Tom Brown, Miljan Martic, Shane Legg, and Dario
  Amodei. 2017.
\newblock Deep reinforcement learning from human preferences.
\newblock \emph{Advances in neural information processing systems}, 30.

\bibitem[{Dong et~al.(2023)Dong, Xiong, Goyal, Zhang, Chow, Pan, Diao, Zhang,
  Shum, and Zhang}]{dong2023raft}
Hanze Dong, Wei Xiong, Deepanshu Goyal, Yihan Zhang, Winnie Chow, Rui Pan,
  Shizhe Diao, Jipeng Zhang, Kashun Shum, and Tong Zhang. 2023.
\newblock \href {https://arxiv.org/abs/2304.06767} {Raft: Reward ranked
  finetuning for generative foundation model alignment}.
\newblock \emph{Preprint}, arXiv:2304.06767.

\bibitem[{Ethayarajh et~al.(2022)Ethayarajh, Choi, and
  Swayamdipta}]{ethayarajh2022understanding}
Kawin Ethayarajh, Yejin Choi, and Swabha Swayamdipta. 2022.
\newblock \href {https://proceedings.mlr.press/v162/ethayarajh22a.html}
  {Understanding dataset difficulty with $\mathcal{V}$-usable information}.
\newblock In \emph{Proceedings of the 39th International Conference on Machine
  Learning}, volume 162 of \emph{Proceedings of Machine Learning Research},
  pages 5988--6008. PMLR.

\bibitem[{Everitt and Hutter(2016)}]{everitt2016avoiding}
Tom Everitt and Marcus Hutter. 2016.
\newblock Avoiding wireheading with value reinforcement learning.
\newblock In \emph{Artificial General Intelligence: 9th International
  Conference, AGI 2016, New York, NY, USA, July 16-19, 2016, Proceedings 9},
  pages 12--22. Springer.

\bibitem[{Fan et~al.(2019)Fan, Jernite, Perez, Grangier, Weston, and
  Auli}]{fan2019eli5}
Angela Fan, Yacine Jernite, Ethan Perez, David Grangier, Jason Weston, and
  Michael Auli. 2019.
\newblock \href {https://arxiv.org/abs/1907.09190} {Eli5: Long form question
  answering}.
\newblock \emph{Preprint}, arXiv:1907.09190.

\bibitem[{Fox et~al.(2015)Fox, Pakman, and Tishby}]{fox2015taming}
Roy Fox, Ari Pakman, and Naftali Tishby. 2015.
\newblock Taming the noise in reinforcement learning via soft updates.
\newblock \emph{arXiv preprint arXiv:1512.08562}.

\bibitem[{Friedman et~al.(2023)Friedman, Ahuja, Allen, Tan, Sidahmed, Long,
  Xie, Schubiner, Patel, Lara et~al.}]{friedman2023leveraging}
Luke Friedman, Sameer Ahuja, David Allen, Terry Tan, Hakim Sidahmed, Changbo
  Long, Jun Xie, Gabriel Schubiner, Ajay Patel, Harsh Lara, et~al. 2023.
\newblock Leveraging large language models in conversational recommender
  systems.
\newblock \emph{arXiv preprint arXiv:2305.07961}.

\bibitem[{Geist et~al.(2019)Geist, Scherrer, and Pietquin}]{geist2019theory}
Matthieu Geist, Bruno Scherrer, and Olivier Pietquin. 2019.
\newblock A theory of regularized markov decision processes.
\newblock In \emph{International Conference on Machine Learning}, pages
  2160--2169. PMLR.

\bibitem[{Glaese et~al.(2022)Glaese, McAleese, Tr{\k{e}}bacz, Aslanides,
  Firoiu, Ewalds, Rauh, Weidinger, Chadwick, Thacker
  et~al.}]{glaese2022improving}
Amelia Glaese, Nat McAleese, Maja Tr{\k{e}}bacz, John Aslanides, Vlad Firoiu,
  Timo Ewalds, Maribeth Rauh, Laura Weidinger, Martin Chadwick, Phoebe Thacker,
  et~al. 2022.
\newblock Improving alignment of dialogue agents via targeted human judgements.
\newblock \emph{arXiv preprint arXiv:2209.14375}.

\bibitem[{Goyal et~al.(2017{\natexlab{a}})Goyal, Khot, Summers-Stay, Batra, and
  Parikh}]{goyal2017making}
Yash Goyal, Tejas Khot, Douglas Summers-Stay, Dhruv Batra, and Devi Parikh.
  2017{\natexlab{a}}.
\newblock Making the v in vqa matter: Elevating the role of image understanding
  in visual question answering.
\newblock In \emph{Proceedings of the IEEE conference on computer vision and
  pattern recognition}, pages 6904--6913.

\bibitem[{Goyal et~al.(2017{\natexlab{b}})Goyal, Khot, Summers{-}Stay, Batra,
  and Parikh}]{balanced_vqa_v2}
Yash Goyal, Tejas Khot, Douglas Summers{-}Stay, Dhruv Batra, and Devi Parikh.
  2017{\natexlab{b}}.
\newblock Making the {V} in {VQA} matter: Elevating the role of image
  understanding in {V}isual {Q}uestion {A}nswering.
\newblock In \emph{Conference on Computer Vision and Pattern Recognition
  (CVPR)}.

\bibitem[{Hu et~al.(2021)Hu, Shen, Wallis, Allen-Zhu, Li, Wang, Wang, and
  Chen}]{hu2021lora}
Edward~J. Hu, Yelong Shen, Phillip Wallis, Zeyuan Allen-Zhu, Yuanzhi Li, Shean
  Wang, Lu~Wang, and Weizhu Chen. 2021.
\newblock \href {https://arxiv.org/abs/2106.09685} {Lora: Low-rank adaptation
  of large language models}.
\newblock \emph{Preprint}, arXiv:2106.09685.

\bibitem[{Jandaghi et~al.(2023)Jandaghi, Sheng, Bai, Pujara, and
  Sidahmed}]{jandaghi2023faithful}
Pegah Jandaghi, XiangHai Sheng, Xinyi Bai, Jay Pujara, and Hakim Sidahmed.
  2023.
\newblock Faithful persona-based conversational dataset generation with large
  language models.
\newblock \emph{arXiv preprint arXiv:2312.10007}.

\bibitem[{Ji et~al.(2023)Ji, Qiu, Chen, Zhang, Lou, Wang, Duan, He, Zhou, Zhang
  et~al.}]{ji2023ai}
Jiaming Ji, Tianyi Qiu, Boyuan Chen, Borong Zhang, Hantao Lou, Kaile Wang,
  Yawen Duan, Zhonghao He, Jiayi Zhou, Zhaowei Zhang, et~al. 2023.
\newblock Ai alignment: A comprehensive survey.
\newblock \emph{arXiv preprint arXiv:2310.19852}.

\bibitem[{Kingma and Ba(2017)}]{kingma2017adam}
Diederik~P. Kingma and Jimmy Ba. 2017.
\newblock \href {https://arxiv.org/abs/1412.6980} {Adam: A method for
  stochastic optimization}.
\newblock \emph{Preprint}, arXiv:1412.6980.

\bibitem[{Lai et~al.(2023)Lai, Nguyen, Ngo, Nguyen, Dernoncourt, Rossi, and
  Nguyen}]{lai2023okapi}
Viet~Dac Lai, Chien~Van Nguyen, Nghia~Trung Ngo, Thuat Nguyen, Franck
  Dernoncourt, Ryan~A. Rossi, and Thien~Huu Nguyen. 2023.
\newblock \href {https://arxiv.org/abs/2307.16039} {Okapi: Instruction-tuned
  large language models in multiple languages with reinforcement learning from
  human feedback}.
\newblock \emph{Preprint}, arXiv:2307.16039.

\bibitem[{Lee et~al.(2023{\natexlab{a}})Lee, Phatale, Mansoor, Lu, Mesnard,
  Bishop, Carbune, and Rastogi}]{lee2023rlaif}
Harrison Lee, Samrat Phatale, Hassan Mansoor, Kellie Lu, Thomas Mesnard, Colton
  Bishop, Victor Carbune, and Abhinav Rastogi. 2023{\natexlab{a}}.
\newblock Rlaif: Scaling reinforcement learning from human feedback with ai
  feedback.
\newblock \emph{arXiv preprint arXiv:2309.00267}.

\bibitem[{Lee et~al.(2023{\natexlab{b}})Lee, Liu, Ryu, Watkins, Du, Boutilier,
  Abbeel, Ghavamzadeh, and Gu}]{lee2023aligning}
Kimin Lee, Hao Liu, Moonkyung Ryu, Olivia Watkins, Yuqing Du, Craig Boutilier,
  Pieter Abbeel, Mohammad Ghavamzadeh, and Shixiang~Shane Gu.
  2023{\natexlab{b}}.
\newblock Aligning text-to-image models using human feedback.
\newblock \emph{arXiv preprint arXiv:2302.12192}.

\bibitem[{Leike et~al.(2018)Leike, Krueger, Everitt, Martic, Maini, and
  Legg}]{leike2018scalable}
Jan Leike, David Krueger, Tom Everitt, Miljan Martic, Vishal Maini, and Shane
  Legg. 2018.
\newblock Scalable agent alignment via reward modeling: a research direction.
\newblock \emph{arXiv preprint arXiv:1811.07871}.

\bibitem[{Li(2021)}]{Li2021macro_action}
Wei Li. 2021.
\newblock \href {https://arxiv.org/abs/2110.08653} {Learning {UI} navigation
  through demonstrations composed of macro actions}.
\newblock \emph{Preprint}, arXiv:2110.08653.

\bibitem[{Li et~al.(2024{\natexlab{a}})Li, Bishop, Li, Rawles, Campbell-Ajala,
  Tyamagundlu, and Riva}]{li2024effects}
Wei Li, William Bishop, Alice Li, Chris Rawles, Folawiyo Campbell-Ajala, Divya
  Tyamagundlu, and Oriana Riva. 2024{\natexlab{a}}.
\newblock On the effects of data scale on computer control agents.
\newblock \emph{arXiv preprint arXiv:2406.03679}.

\bibitem[{Li et~al.(2024{\natexlab{b}})Li, Bishop, Li, Rawles, Campbell-Ajala,
  Tyamagundlu, and Riva}]{AndroidControl}
Wei Li, William Bishop, Alice Li, Chris Rawles, Folawiyo Campbell-Ajala, Divya
  Tyamagundlu, and Oriana Riva. 2024{\natexlab{b}}.
\newblock \href {https://arxiv.org/abs/2406.03679} {On the effects of data
  scale on computer control agents}.
\newblock \emph{Preprint}, arXiv:2406.03679.

\bibitem[{Li et~al.(2024{\natexlab{c}})Li, Hsu, Bishop, Campbell-Ajala, Lin,
  and Riva}]{li2024uinav}
Wei Li, Fu-Lin Hsu, Will Bishop, Folawiyo Campbell-Ajala, Max Lin, and Oriana
  Riva. 2024{\natexlab{c}}.
\newblock {UINav}: A practical approach to train {On-Device} automation agents.
\newblock In \emph{{NAACL} Industry}.

\bibitem[{Liu et~al.(2024)Liu, Wang, Yin, Molchanov, Wang, Cheng, and
  Chen}]{liu2024dora}
Shih-Yang Liu, Chien-Yi Wang, Hongxu Yin, Pavlo Molchanov, Yu-Chiang~Frank
  Wang, Kwang-Ting Cheng, and Min-Hung Chen. 2024.
\newblock Dora: Weight-decomposed low-rank adaptation.
\newblock \emph{arXiv preprint arXiv:2402.09353}.

\bibitem[{OpenAI et~al.(2023)OpenAI, :, Achiam, Adler, Agarwal, Ahmad, Akkaya,
  Aleman, Almeida, Altenschmidt, Altman, Anadkat, Avila, Babuschkin, Balaji,
  Balcom, Baltescu, Bao, Bavarian, Belgum, Bello, Berdine, Bernadett-Shapiro,
  Berner, Bogdonoff, Boiko, Boyd, Brakman, Brockman, Brooks, Brundage, Button,
  Cai, Campbell, Cann, Carey, Carlson, Carmichael, Chan, Chang, Chantzis, Chen,
  Chen, Chen, Chen, Chen, Chess, Cho, Chu, Chung, Cummings, Currier, Dai,
  Decareaux, Degry, Deutsch, Deville, Dhar, Dohan, Dowling, Dunning, Ecoffet,
  Eleti, Eloundou, Farhi, Fedus, Felix, Fishman, Forte, Fulford, Gao, Georges,
  Gibson, Goel, Gogineni, Goh, Gontijo-Lopes, Gordon, Grafstein, Gray, Greene,
  Gross, Gu, Guo, Hallacy, Han, Harris, He, Heaton, Heidecke, Hesse, Hickey,
  Hickey, Hoeschele, Houghton, Hsu, Hu, Hu, Huizinga, Jain, Jain, Jang, Jiang,
  Jiang, Jin, Jin, Jomoto, Jonn, Jun, Kaftan, Łukasz Kaiser, Kamali,
  Kanitscheider, Keskar, Khan, Kilpatrick, Kim, Kim, Kim, Kirchner, Kiros,
  Knight, Kokotajlo, Łukasz Kondraciuk, Kondrich, Konstantinidis, Kosic,
  Krueger, Kuo, Lampe, Lan, Lee, Leike, Leung, Levy, Li, Lim, Lin, Lin, Litwin,
  Lopez, Lowe, Lue, Makanju, Malfacini, Manning, Markov, Markovski, Martin,
  Mayer, Mayne, McGrew, McKinney, McLeavey, McMillan, McNeil, Medina, Mehta,
  Menick, Metz, Mishchenko, Mishkin, Monaco, Morikawa, Mossing, Mu, Murati,
  Murk, Mély, Nair, Nakano, Nayak, Neelakantan, Ngo, Noh, Ouyang, O'Keefe,
  Pachocki, Paino, Palermo, Pantuliano, Parascandolo, Parish, Parparita,
  Passos, Pavlov, Peng, Perelman, de~Avila Belbute~Peres, Petrov,
  de~Oliveira~Pinto, Michael, Pokorny, Pokrass, Pong, Powell, Power, Power,
  Proehl, Puri, Radford, Rae, Ramesh, Raymond, Real, Rimbach, Ross, Rotsted,
  Roussez, Ryder, Saltarelli, Sanders, Santurkar, Sastry, Schmidt, Schnurr,
  Schulman, Selsam, Sheppard, Sherbakov, Shieh, Shoker, Shyam, Sidor, Sigler,
  Simens, Sitkin, Slama, Sohl, Sokolowsky, Song, Staudacher, Such, Summers,
  Sutskever, Tang, Tezak, Thompson, Tillet, Tootoonchian, Tseng, Tuggle,
  Turley, Tworek, Uribe, Vallone, Vijayvergiya, Voss, Wainwright, Wang, Wang,
  Wang, Ward, Wei, Weinmann, Welihinda, Welinder, Weng, Weng, Wiethoff,
  Willner, Winter, Wolrich, Wong, Workman, Wu, Wu, Wu, Xiao, Xu, Yoo, Yu, Yuan,
  Zaremba, Zellers, Zhang, Zhang, Zhao, Zheng, Zhuang, Zhuk, and
  Zoph}]{openai2023gpt4}
OpenAI, :, Josh Achiam, Steven Adler, Sandhini Agarwal, Lama Ahmad, Ilge
  Akkaya, Florencia~Leoni Aleman, Diogo Almeida, Janko Altenschmidt, Sam
  Altman, Shyamal Anadkat, Red Avila, Igor Babuschkin, Suchir Balaji, Valerie
  Balcom, Paul Baltescu, Haiming Bao, Mo~Bavarian, Jeff Belgum, Irwan Bello,
  Jake Berdine, Gabriel Bernadett-Shapiro, Christopher Berner, Lenny Bogdonoff,
  Oleg Boiko, Madelaine Boyd, Anna-Luisa Brakman, Greg Brockman, Tim Brooks,
  Miles Brundage, Kevin Button, Trevor Cai, Rosie Campbell, Andrew Cann,
  Brittany Carey, Chelsea Carlson, Rory Carmichael, Brooke Chan, Che Chang,
  Fotis Chantzis, Derek Chen, Sully Chen, Ruby Chen, Jason Chen, Mark Chen, Ben
  Chess, Chester Cho, Casey Chu, Hyung~Won Chung, Dave Cummings, Jeremiah
  Currier, Yunxing Dai, Cory Decareaux, Thomas Degry, Noah Deutsch, Damien
  Deville, Arka Dhar, David Dohan, Steve Dowling, Sheila Dunning, Adrien
  Ecoffet, Atty Eleti, Tyna Eloundou, David Farhi, Liam Fedus, Niko Felix,
  Simón~Posada Fishman, Juston Forte, Isabella Fulford, Leo Gao, Elie Georges,
  Christian Gibson, Vik Goel, Tarun Gogineni, Gabriel Goh, Rapha Gontijo-Lopes,
  Jonathan Gordon, Morgan Grafstein, Scott Gray, Ryan Greene, Joshua Gross,
  Shixiang~Shane Gu, Yufei Guo, Chris Hallacy, Jesse Han, Jeff Harris, Yuchen
  He, Mike Heaton, Johannes Heidecke, Chris Hesse, Alan Hickey, Wade Hickey,
  Peter Hoeschele, Brandon Houghton, Kenny Hsu, Shengli Hu, Xin Hu, Joost
  Huizinga, Shantanu Jain, Shawn Jain, Joanne Jang, Angela Jiang, Roger Jiang,
  Haozhun Jin, Denny Jin, Shino Jomoto, Billie Jonn, Heewoo Jun, Tomer Kaftan,
  Łukasz Kaiser, Ali Kamali, Ingmar Kanitscheider, Nitish~Shirish Keskar,
  Tabarak Khan, Logan Kilpatrick, Jong~Wook Kim, Christina Kim, Yongjik Kim,
  Hendrik Kirchner, Jamie Kiros, Matt Knight, Daniel Kokotajlo, Łukasz
  Kondraciuk, Andrew Kondrich, Aris Konstantinidis, Kyle Kosic, Gretchen
  Krueger, Vishal Kuo, Michael Lampe, Ikai Lan, Teddy Lee, Jan Leike, Jade
  Leung, Daniel Levy, Chak~Ming Li, Rachel Lim, Molly Lin, Stephanie Lin,
  Mateusz Litwin, Theresa Lopez, Ryan Lowe, Patricia Lue, Anna Makanju, Kim
  Malfacini, Sam Manning, Todor Markov, Yaniv Markovski, Bianca Martin, Katie
  Mayer, Andrew Mayne, Bob McGrew, Scott~Mayer McKinney, Christine McLeavey,
  Paul McMillan, Jake McNeil, David Medina, Aalok Mehta, Jacob Menick, Luke
  Metz, Andrey Mishchenko, Pamela Mishkin, Vinnie Monaco, Evan Morikawa, Daniel
  Mossing, Tong Mu, Mira Murati, Oleg Murk, David Mély, Ashvin Nair, Reiichiro
  Nakano, Rajeev Nayak, Arvind Neelakantan, Richard Ngo, Hyeonwoo Noh, Long
  Ouyang, Cullen O'Keefe, Jakub Pachocki, Alex Paino, Joe Palermo, Ashley
  Pantuliano, Giambattista Parascandolo, Joel Parish, Emy Parparita, Alex
  Passos, Mikhail Pavlov, Andrew Peng, Adam Perelman, Filipe de~Avila
  Belbute~Peres, Michael Petrov, Henrique~Ponde de~Oliveira~Pinto, Michael,
  Pokorny, Michelle Pokrass, Vitchyr Pong, Tolly Powell, Alethea Power, Boris
  Power, Elizabeth Proehl, Raul Puri, Alec Radford, Jack Rae, Aditya Ramesh,
  Cameron Raymond, Francis Real, Kendra Rimbach, Carl Ross, Bob Rotsted, Henri
  Roussez, Nick Ryder, Mario Saltarelli, Ted Sanders, Shibani Santurkar, Girish
  Sastry, Heather Schmidt, David Schnurr, John Schulman, Daniel Selsam, Kyla
  Sheppard, Toki Sherbakov, Jessica Shieh, Sarah Shoker, Pranav Shyam, Szymon
  Sidor, Eric Sigler, Maddie Simens, Jordan Sitkin, Katarina Slama, Ian Sohl,
  Benjamin Sokolowsky, Yang Song, Natalie Staudacher, Felipe~Petroski Such,
  Natalie Summers, Ilya Sutskever, Jie Tang, Nikolas Tezak, Madeleine Thompson,
  Phil Tillet, Amin Tootoonchian, Elizabeth Tseng, Preston Tuggle, Nick Turley,
  Jerry Tworek, Juan Felipe~Cerón Uribe, Andrea Vallone, Arun Vijayvergiya,
  Chelsea Voss, Carroll Wainwright, Justin~Jay Wang, Alvin Wang, Ben Wang,
  Jonathan Ward, Jason Wei, CJ~Weinmann, Akila Welihinda, Peter Welinder, Jiayi
  Weng, Lilian Weng, Matt Wiethoff, Dave Willner, Clemens Winter, Samuel
  Wolrich, Hannah Wong, Lauren Workman, Sherwin Wu, Jeff Wu, Michael Wu, Kai
  Xiao, Tao Xu, Sarah Yoo, Kevin Yu, Qiming Yuan, Wojciech Zaremba, Rowan
  Zellers, Chong Zhang, Marvin Zhang, Shengjia Zhao, Tianhao Zheng, Juntang
  Zhuang, William Zhuk, and Barret Zoph. 2023.
\newblock \href {https://arxiv.org/abs/2303.08774} {Gpt-4 technical report}.
\newblock \emph{Preprint}, arXiv:2303.08774.

\bibitem[{Ouyang et~al.(2022)Ouyang, Wu, Jiang, Almeida, Wainwright, Mishkin,
  Zhang, Agarwal, Slama, Ray et~al.}]{ouyang2022training}
Long Ouyang, Jeffrey Wu, Xu~Jiang, Diogo Almeida, Carroll Wainwright, Pamela
  Mishkin, Chong Zhang, Sandhini Agarwal, Katarina Slama, Alex Ray, et~al.
  2022.
\newblock Training language models to follow instructions with human feedback.
\newblock \emph{Advances in Neural Information Processing Systems},
  35:27730--27744.

\bibitem[{Paxml(2022)}]{paxml}
Paxml. 2022.
\newblock \emph{Paxml: a Jax-based machine learning framework for training
  large scale models}.
\newblock \url{https://github.com/google/paxml} [Accessed: 2024-01-03].

\bibitem[{Radford et~al.(2018)Radford, Narasimhan, Salimans, Sutskever
  et~al.}]{radford2018improving}
Alec Radford, Karthik Narasimhan, Tim Salimans, Ilya Sutskever, et~al. 2018.
\newblock Improving language understanding by generative pre-training.

\bibitem[{Rafailov et~al.(2023)Rafailov, Sharma, Mitchell, Ermon, Manning, and
  Finn}]{rafailov2023direct}
Rafael Rafailov, Archit Sharma, Eric Mitchell, Stefano Ermon, Christopher~D
  Manning, and Chelsea Finn. 2023.
\newblock Direct preference optimization: Your language model is secretly a
  reward model.
\newblock \emph{arXiv preprint arXiv:2305.18290}.

\bibitem[{Ramachandran et~al.(2016)Ramachandran, Liu, and
  Le}]{ramachandran2016unsupervised}
Prajit Ramachandran, Peter~J Liu, and Quoc~V Le. 2016.
\newblock Unsupervised pretraining for sequence to sequence learning.
\newblock \emph{arXiv preprint arXiv:1611.02683}.

\bibitem[{Ram{\'e} et~al.(2024)Ram{\'e}, Vieillard, Hussenot, Dadashi, Cideron,
  Bachem, and Ferret}]{rame2024warm}
Alexandre Ram{\'e}, Nino Vieillard, L{\'e}onard Hussenot, Robert Dadashi,
  Geoffrey Cideron, Olivier Bachem, and Johan Ferret. 2024.
\newblock Warm: On the benefits of weight averaged reward models.
\newblock \emph{arXiv preprint arXiv:2401.12187}.

\bibitem[{Reid et~al.(2024)Reid, Savinov, Teplyashin, Lepikhin, Lillicrap,
  Alayrac, Soricut, Lazaridou, Firat, Schrittwieser et~al.}]{reid2024gemini}
Machel Reid, Nikolay Savinov, Denis Teplyashin, Dmitry Lepikhin, Timothy
  Lillicrap, Jean-baptiste Alayrac, Radu Soricut, Angeliki Lazaridou, Orhan
  Firat, Julian Schrittwieser, et~al. 2024.
\newblock Gemini 1.5: Unlocking multimodal understanding across millions of
  tokens of context.
\newblock \emph{arXiv preprint arXiv:2403.05530}.

\bibitem[{Roberts et~al.(2022)Roberts, Chung, Levskaya, Mishra, Bradbury,
  Andor, Narang, Lester, Gaffney, Mohiuddin, Hawthorne, Lewkowycz, Salcianu,
  van Zee, Austin, Goodman, Soares, Hu, Tsvyashchenko, Chowdhery, Bastings,
  Bulian, Garcia, Ni, Chen, Kenealy, Clark, Lee, Garrette, Lee-Thorp, Raffel,
  Shazeer, Ritter, Bosma, Passos, Maitin-Shepard, Fiedel, Omernick, Saeta,
  Sepassi, Spiridonov, Newlan, and Gesmundo}]{roberts2022scaling}
Adam Roberts, Hyung~Won Chung, Anselm Levskaya, Gaurav Mishra, James Bradbury,
  Daniel Andor, Sharan Narang, Brian Lester, Colin Gaffney, Afroz Mohiuddin,
  Curtis Hawthorne, Aitor Lewkowycz, Alex Salcianu, Marc van Zee, Jacob Austin,
  Sebastian Goodman, Livio~Baldini Soares, Haitang Hu, Sasha Tsvyashchenko,
  Aakanksha Chowdhery, Jasmijn Bastings, Jannis Bulian, Xavier Garcia, Jianmo
  Ni, Andrew Chen, Kathleen Kenealy, Jonathan~H. Clark, Stephan Lee, Dan
  Garrette, James Lee-Thorp, Colin Raffel, Noam Shazeer, Marvin Ritter, Maarten
  Bosma, Alexandre Passos, Jeremy Maitin-Shepard, Noah Fiedel, Mark Omernick,
  Brennan Saeta, Ryan Sepassi, Alexander Spiridonov, Joshua Newlan, and Andrea
  Gesmundo. 2022.
\newblock \href {https://arxiv.org/abs/2203.17189} {Scaling up models and data
  with $\texttt{t5x}$ and $\texttt{seqio}$}.
\newblock \emph{Preprint}, arXiv:2203.17189.

\bibitem[{Shazeer and Stern(2018)}]{shazeer2018adafactor}
Noam Shazeer and Mitchell Stern. 2018.
\newblock \href {https://arxiv.org/abs/1804.04235} {Adafactor: Adaptive
  learning rates with sublinear memory cost}.
\newblock \emph{Preprint}, arXiv:1804.04235.

\bibitem[{Stiennon et~al.(2020)Stiennon, Ouyang, Wu, Ziegler, Lowe, Voss,
  Radford, Amodei, and Christiano}]{stiennon2020learning}
Nisan Stiennon, Long Ouyang, Jeffrey Wu, Daniel Ziegler, Ryan Lowe, Chelsea
  Voss, Alec Radford, Dario Amodei, and Paul~F Christiano. 2020.
\newblock Learning to summarize with human feedback.
\newblock \emph{Advances in Neural Information Processing Systems},
  33:3008--3021.

\bibitem[{Sun et~al.(2023)Sun, Shen, Cao, Liu, Li, Shen, Gan, Gui, Wang, Yang
  et~al.}]{sun2023aligning}
Zhiqing Sun, Sheng Shen, Shengcao Cao, Haotian Liu, Chunyuan Li, Yikang Shen,
  Chuang Gan, Liang-Yan Gui, Yu-Xiong Wang, Yiming Yang, et~al. 2023.
\newblock Aligning large multimodal models with factually augmented rlhf.
\newblock \emph{arXiv preprint arXiv:2309.14525}.

\bibitem[{Tay et~al.(2022)Tay, Dehghani, Tran, Garcia, Wei, Wang, Chung,
  Shakeri, Bahri, Schuster et~al.}]{tay2022ul2}
Yi~Tay, Mostafa Dehghani, Vinh~Q Tran, Xavier Garcia, Jason Wei, Xuezhi Wang,
  Hyung~Won Chung, Siamak Shakeri, Dara Bahri, Tal Schuster, et~al. 2022.
\newblock Ul2: Unifying language learning paradigms.
\newblock \emph{arXiv preprint arXiv:2205.05131}.

\bibitem[{Team et~al.(2023)Team, Anil, Borgeaud, Wu, Alayrac, Yu, Soricut,
  Schalkwyk, Dai, Hauth, Millican, Silver, Petrov, Johnson, Antonoglou,
  Schrittwieser, Glaese, Chen, Pitler, Lillicrap, Lazaridou, Firat, Molloy,
  Isard, Barham, Hennigan, Lee, Viola, Reynolds, Xu, Doherty, Collins, Meyer,
  Rutherford, Moreira, Ayoub, Goel, Tucker, Piqueras, Krikun, Barr, Savinov,
  Danihelka, Roelofs, White, Andreassen, von Glehn, Yagati, Kazemi, Gonzalez,
  Khalman, Sygnowski, Frechette, Smith, Culp, Proleev, Luan, Chen, Lottes,
  Schucher, Lebron, Rrustemi, Clay, Crone, Kocisky, Zhao, Perz, Yu, Howard,
  Bloniarz, Rae, Lu, Sifre, Maggioni, Alcober, Garrette, Barnes, Thakoor,
  Austin, Barth-Maron, Wong, Joshi, Chaabouni, Fatiha, Ahuja, Liu, Li, Cogan,
  Chen, Jia, Gu, Zhang, Grimstad, Hartman, Chadwick, Tomar, Garcia, Senter,
  Taropa, Pillai, Devlin, Laskin, de~Las~Casas, Valter, Tao, Blanco, Badia,
  Reitter, Chen, Brennan, Rivera, Brin, Iqbal, Surita, Labanowski, Rao,
  Winkler, Parisotto, Gu, Olszewska, Zhang, Addanki, Miech, Louis, Shafey,
  Teplyashin, Brown, Catt, Attaluri, Balaguer, Xiang, Wang, Ashwood, Briukhov,
  Webson, Ganapathy, Sanghavi, Kannan, Chang, Stjerngren, Djolonga, Sun, Bapna,
  Aitchison, Pejman, Michalewski, Yu, Wang, Love, Ahn, Bloxwich, Han,
  Humphreys, Sellam, Bradbury, Godbole, Samangooei, Damoc, Kaskasoli, Arnold,
  Vasudevan, Agrawal, Riesa, Lepikhin, Tanburn, Srinivasan, Lim, Hodkinson,
  Shyam, Ferret, Hand, Garg, Paine, Li, Li, Giang, Neitz, Abbas, York, Reid,
  Cole, Chowdhery, Das, Rogozińska, Nikolaev, Sprechmann, Nado, Zilka, Prost,
  He, Monteiro, Mishra, Welty, Newlan, Jia, Allamanis, Hu, de~Liedekerke,
  Gilmer, Saroufim, Rijhwani, Hou, Shrivastava, Baddepudi, Goldin, Ozturel,
  Cassirer, Xu, Sohn, Sachan, Amplayo, Swanson, Petrova, Narayan, Guez, Brahma,
  Landon, Patel, Zhao, Villela, Wang, Jia, Rahtz, Giménez, Yeung, Lin,
  Keeling, Georgiev, Mincu, Wu, Haykal, Saputro, Vodrahalli, Qin, Cankara,
  Sharma, Fernando, Hawkins, Neyshabur, Kim, Hutter, Agrawal, Castro-Ros,
  van~den Driessche, Wang, Yang, yiin Chang, Komarek, McIlroy, Lučić, Zhang,
  Farhan, Sharman, Natsev, Michel, Cheng, Bansal, Qiao, Cao, Shakeri,
  Butterfield, Chung, Rubenstein, Agrawal, Mensch, Soparkar, Lenc, Chung, Pope,
  Maggiore, Kay, Jhakra, Wang, Maynez, Phuong, Tobin, Tacchetti, Trebacz,
  Robinson, Katariya, Riedel, Bailey, Xiao, Ghelani, Aroyo, Slone, Houlsby,
  Xiong, Yang, Gribovskaya, Adler, Wirth, Lee, Li, Kagohara, Pavagadhi,
  Bridgers, Bortsova, Ghemawat, Ahmed, Liu, Powell, Bolina, Iinuma,
  Zablotskaia, Besley, Chung, Dozat, Comanescu, Si, Greer, Su, Polacek,
  Kaufman, Tokumine, Hu, Buchatskaya, Miao, Elhawaty, Siddhant, Tomasev, Xing,
  Greer, Miller, Ashraf, Roy, Zhang, Ma, Filos, Besta, Blevins, Klimenko, Yeh,
  Changpinyo, Mu, Chang, Pajarskas, Muir, Cohen, Lan, Haridasan, Marathe,
  Hansen, Douglas, Samuel, Wang, Austin, Lan, Jiang, Chiu, Lorenzo, Sjösund,
  Cevey, Gleicher, Avrahami, Boral, Srinivasan, Selo, May, Aisopos, Hussenot,
  Soares, Baumli, Chang, Recasens, Caine, Pritzel, Pavetic, Pardo, Gergely,
  Frye, Ramasesh, Horgan, Badola, Kassner, Roy, Dyer, Campos, Tomala, Tang,
  Badawy, White, Mustafa, Lang, Jindal, Vikram, Gong, Caelles, Hemsley,
  Thornton, Feng, Stokowiec, Zheng, Thacker, Çağlar Ünlü, Zhang, Saleh,
  Svensson, Bileschi, Patil, Anand, Ring, Tsihlas, Vezer, Selvi, Shevlane,
  Rodriguez, Kwiatkowski, Daruki, Rong, Dafoe, FitzGerald, Gu-Lemberg, Khan,
  Hendricks, Pellat, Feinberg, Cobon-Kerr, Sainath, Rauh, Hashemi, Ives,
  Hasson, Li, Noland, Cao, Byrd, Hou, Wang, Sottiaux, Paganini, Lespiau,
  Moufarek, Hassan, Shivakumar, van Amersfoort, Mandhane, Joshi, Goyal, Tung,
  Brock, Sheahan, Misra, Li, Rakićević, Dehghani, Liu, Mittal, Oh, Noury,
  Sezener, Huot, Lamm, Cao, Chen, Elsayed, Chi, Mahdieh, Tenney, Hua,
  Petrychenko, Kane, Scandinaro, Jain, Uesato, Datta, Sadovsky, Bunyan, Rabiej,
  Wu, Zhang, Vasudevan, Leurent, Alnahlawi, Georgescu, Wei, Zheng, Chan,
  Rabinovitch, Stanczyk, Zhang, Steiner, Naskar, Azzam, Johnson, Paszke, Chiu,
  Elias, Mohiuddin, Muhammad, Miao, Lee, Vieillard, Potluri, Park, Davoodi,
  Zhang, Stanway, Garmon, Karmarkar, Dong, Lee, Kumar, Zhou, Evens, Isaac,
  Chen, Jia, Levskaya, Zhu, Gorgolewski, Grabowski, Mao, Magni, Yao, Snaider,
  Casagrande, Suganthan, Palmer, Irving, Loper, Faruqui, Arkatkar, Chen,
  Shafran, Fink, Castaño, Giannoumis, Kim, Rybiński, Sreevatsa, Prendki,
  Soergel, Goedeckemeyer, Gierke, Jafari, Gaba, Wiesner, Wright, Wei, Vashisht,
  Kulizhskaya, Hoover, Le, Li, Iwuanyanwu, Liu, Ramirez, Khorlin, Cui, LIN,
  Georgiev, Wu, Aguilar, Pallo, Chakladar, Repina, Wu, van~der Weide,
  Ponnapalli, Kaplan, Simsa, Li, Dousse, Yang, Piper, Ie, Lui, Pasumarthi,
  Lintz, Vijayakumar, Thiet, Andor, Valenzuela, Paduraru, Peng, Lee, Zhang,
  Greene, Nguyen, Kurylowicz, Velury, Krause, Hardin, Dixon, Janzer, Choo,
  Feng, Zhang, Singhal, Latkar, Zhang, Le, Abellan, Du, McKinnon, Antropova,
  Bolukbasi, Keller, Reid, Finchelstein, Raad, Crocker, Hawkins, Dadashi,
  Gaffney, Lall, Franko, Filonov, Bulanova, Leblond, Yadav, Chung, Askham,
  Cobo, Xu, Fischer, Xu, Sorokin, Alberti, Lin, Evans, Zhou, Dimitriev, Forbes,
  Banarse, Tung, Liu, Omernick, Bishop, Kumar, Sterneck, Foley, Jain, Mishra,
  Xia, Bos, Cideron, Amid, Piccinno, Wang, Banzal, Gurita, Noga, Shah,
  Mankowitz, Polozov, Kushman, Krakovna, Brown, Bateni, Duan, Firoiu,
  Thotakuri, Natan, Mohananey, Geist, Mudgal, Girgin, Li, Ye, Roval, Tojo,
  Kwong, Lee-Thorp, Yew, Yuan, Bagri, Sinopalnikov, Ramos, Mellor, Sharma,
  Severyn, Lai, Wu, Cheng, Miller, Sonnerat, Vnukov, Greig, Beattie, Caveness,
  Bai, Eisenschlos, Korchemniy, Tsai, Jasarevic, Kong, Dao, Zheng, Liu, Yang,
  Zhu, Geller, Teh, Sanmiya, Gladchenko, Trdin, Sozanschi, Toyama, Rosen,
  Tavakkol, Xue, Elkind, Woodman, Carpenter, Papamakarios, Kemp, Kafle,
  Grunina, Sinha, Talbert, Goyal, Wu, Owusu-Afriyie, Du, Thornton, Pont-Tuset,
  Narayana, Li, Fatehi, Wieting, Ajmeri, Uria, Zhu, Ko, Knight, Héliou, Niu,
  Gu, Pang, Tran, Li, Levine, Stolovich, Kalb, Santamaria-Fernandez, Goenka,
  Yustalim, Strudel, Elqursh, Lakshminarayanan, Deck, Upadhyay, Lee,
  Dusenberry, Li, Wang, Levin, Hoffmann, Holtmann-Rice, Bachem, Yue, Arora,
  Malmi, Mirylenka, Tan, Koh, Yeganeh, Põder, Zheng, Pongetti, Tariq, Sun,
  Ionita, Seyedhosseini, Tafti, Kotikalapudi, Liu, Gulati, Liu, Ye, Chrzaszcz,
  Wang, Sethi, Li, Brown, Singh, Fan, Parisi, Stanton, Kuang, Koverkathu,
  Choquette-Choo, Li, Lu, Ittycheriah, Shroff, Sun, Varadarajan, Bahargam,
  Willoughby, Gaddy, Dasgupta, Desjardins, Cornero, Robenek, Mittal, Albrecht,
  Shenoy, Moiseev, Jacobsson, Ghaffarkhah, Rivière, Walton, Crepy, Parrish,
  Liu, Zhou, Farabet, Radebaugh, Srinivasan, van~der Salm, Fidjeland, Scellato,
  Latorre-Chimoto, Klimczak-Plucińska, Bridson, de~Cesare, Hudson,
  Mendolicchio, Walker, Morris, Penchev, Mauger, Guseynov, Reid, Odoom, Loher,
  Cotruta, Yenugula, Grewe, Petrushkina, Duerig, Sanchez, Yadlowsky, Shen,
  Globerson, Kurzrok, Webb, Dua, Li, Lahoti, Bhupatiraju, Hurt, Qureshi,
  Agarwal, Shani, Eyal, Khare, Belle, Wang, Tekur, Kale, Wei, Sang, Saeta,
  Liechty, Sun, Zhao, Lee, Nayak, Fritz, Vuyyuru, Aslanides, Vyas, Wicke, Ma,
  Bilal, Eltyshev, Balle, Martin, Cate, Manyika, Amiri, Kim, Xiong, Kang,
  Luisier, Tripuraneni, Madras, Guo, Waters, Wang, Ainslie, Baldridge, Zhang,
  Pruthi, Bauer, Yang, Mansour, Gelman, Xu, Polovets, Liu, Cai, Chen, Sheng,
  Xue, Ozair, Yu, Angermueller, Li, Wang, Wiesinger, Koukoumidis, Tian, Iyer,
  Gurumurthy, Goldenson, Shah, Blake, Yu, Urbanowicz, Palomaki, Fernando,
  Brooks, Durden, Mehta, Momchev, Rahimtoroghi, Georgaki, Raul, Ruder, Redshaw,
  Lee, Jalan, Li, Perng, Hechtman, Schuh, Nasr, Chen, Milan, Mikulik, Strohman,
  Franco, Green, Hassabis, Kavukcuoglu, Dean, and
  Vinyals}]{geminiteam2023gemini}
Gemini Team, Rohan Anil, Sebastian Borgeaud, Yonghui Wu, Jean-Baptiste Alayrac,
  Jiahui Yu, Radu Soricut, Johan Schalkwyk, Andrew~M. Dai, Anja Hauth, Katie
  Millican, David Silver, Slav Petrov, Melvin Johnson, Ioannis Antonoglou,
  Julian Schrittwieser, Amelia Glaese, Jilin Chen, Emily Pitler, Timothy
  Lillicrap, Angeliki Lazaridou, Orhan Firat, James Molloy, Michael Isard,
  Paul~R. Barham, Tom Hennigan, Benjamin Lee, Fabio Viola, Malcolm Reynolds,
  Yuanzhong Xu, Ryan Doherty, Eli Collins, Clemens Meyer, Eliza Rutherford,
  Erica Moreira, Kareem Ayoub, Megha Goel, George Tucker, Enrique Piqueras,
  Maxim Krikun, Iain Barr, Nikolay Savinov, Ivo Danihelka, Becca Roelofs,
  Anaïs White, Anders Andreassen, Tamara von Glehn, Lakshman Yagati, Mehran
  Kazemi, Lucas Gonzalez, Misha Khalman, Jakub Sygnowski, Alexandre Frechette,
  Charlotte Smith, Laura Culp, Lev Proleev, Yi~Luan, Xi~Chen, James Lottes,
  Nathan Schucher, Federico Lebron, Alban Rrustemi, Natalie Clay, Phil Crone,
  Tomas Kocisky, Jeffrey Zhao, Bartek Perz, Dian Yu, Heidi Howard, Adam
  Bloniarz, Jack~W. Rae, Han Lu, Laurent Sifre, Marcello Maggioni, Fred
  Alcober, Dan Garrette, Megan Barnes, Shantanu Thakoor, Jacob Austin, Gabriel
  Barth-Maron, William Wong, Rishabh Joshi, Rahma Chaabouni, Deeni Fatiha, Arun
  Ahuja, Ruibo Liu, Yunxuan Li, Sarah Cogan, Jeremy Chen, Chao Jia, Chenjie Gu,
  Qiao Zhang, Jordan Grimstad, Ale~Jakse Hartman, Martin Chadwick, Gaurav~Singh
  Tomar, Xavier Garcia, Evan Senter, Emanuel Taropa,
  Thanumalayan~Sankaranarayana Pillai, Jacob Devlin, Michael Laskin, Diego
  de~Las~Casas, Dasha Valter, Connie Tao, Lorenzo Blanco, Adrià~Puigdomènech
  Badia, David Reitter, Mianna Chen, Jenny Brennan, Clara Rivera, Sergey Brin,
  Shariq Iqbal, Gabriela Surita, Jane Labanowski, Abhi Rao, Stephanie Winkler,
  Emilio Parisotto, Yiming Gu, Kate Olszewska, Yujing Zhang, Ravi Addanki,
  Antoine Miech, Annie Louis, Laurent~El Shafey, Denis Teplyashin, Geoff Brown,
  Elliot Catt, Nithya Attaluri, Jan Balaguer, Jackie Xiang, Pidong Wang, Zoe
  Ashwood, Anton Briukhov, Albert Webson, Sanjay Ganapathy, Smit Sanghavi, Ajay
  Kannan, Ming-Wei Chang, Axel Stjerngren, Josip Djolonga, Yuting Sun, Ankur
  Bapna, Matthew Aitchison, Pedram Pejman, Henryk Michalewski, Tianhe Yu, Cindy
  Wang, Juliette Love, Junwhan Ahn, Dawn Bloxwich, Kehang Han, Peter Humphreys,
  Thibault Sellam, James Bradbury, Varun Godbole, Sina Samangooei, Bogdan
  Damoc, Alex Kaskasoli, Sébastien M.~R. Arnold, Vijay Vasudevan, Shubham
  Agrawal, Jason Riesa, Dmitry Lepikhin, Richard Tanburn, Srivatsan Srinivasan,
  Hyeontaek Lim, Sarah Hodkinson, Pranav Shyam, Johan Ferret, Steven Hand,
  Ankush Garg, Tom~Le Paine, Jian Li, Yujia Li, Minh Giang, Alexander Neitz,
  Zaheer Abbas, Sarah York, Machel Reid, Elizabeth Cole, Aakanksha Chowdhery,
  Dipanjan Das, Dominika Rogozińska, Vitaly Nikolaev, Pablo Sprechmann,
  Zachary Nado, Lukas Zilka, Flavien Prost, Luheng He, Marianne Monteiro,
  Gaurav Mishra, Chris Welty, Josh Newlan, Dawei Jia, Miltiadis Allamanis,
  Clara~Huiyi Hu, Raoul de~Liedekerke, Justin Gilmer, Carl Saroufim, Shruti
  Rijhwani, Shaobo Hou, Disha Shrivastava, Anirudh Baddepudi, Alex Goldin,
  Adnan Ozturel, Albin Cassirer, Yunhan Xu, Daniel Sohn, Devendra Sachan,
  Reinald~Kim Amplayo, Craig Swanson, Dessie Petrova, Shashi Narayan, Arthur
  Guez, Siddhartha Brahma, Jessica Landon, Miteyan Patel, Ruizhe Zhao, Kevin
  Villela, Luyu Wang, Wenhao Jia, Matthew Rahtz, Mai Giménez, Legg Yeung,
  Hanzhao Lin, James Keeling, Petko Georgiev, Diana Mincu, Boxi Wu, Salem
  Haykal, Rachel Saputro, Kiran Vodrahalli, James Qin, Zeynep Cankara, Abhanshu
  Sharma, Nick Fernando, Will Hawkins, Behnam Neyshabur, Solomon Kim, Adrian
  Hutter, Priyanka Agrawal, Alex Castro-Ros, George van~den Driessche, Tao
  Wang, Fan Yang, Shuo yiin Chang, Paul Komarek, Ross McIlroy, Mario Lučić,
  Guodong Zhang, Wael Farhan, Michael Sharman, Paul Natsev, Paul Michel, Yong
  Cheng, Yamini Bansal, Siyuan Qiao, Kris Cao, Siamak Shakeri, Christina
  Butterfield, Justin Chung, Paul~Kishan Rubenstein, Shivani Agrawal, Arthur
  Mensch, Kedar Soparkar, Karel Lenc, Timothy Chung, Aedan Pope, Loren
  Maggiore, Jackie Kay, Priya Jhakra, Shibo Wang, Joshua Maynez, Mary Phuong,
  Taylor Tobin, Andrea Tacchetti, Maja Trebacz, Kevin Robinson, Yash Katariya,
  Sebastian Riedel, Paige Bailey, Kefan Xiao, Nimesh Ghelani, Lora Aroyo,
  Ambrose Slone, Neil Houlsby, Xuehan Xiong, Zhen Yang, Elena Gribovskaya,
  Jonas Adler, Mateo Wirth, Lisa Lee, Music Li, Thais Kagohara, Jay Pavagadhi,
  Sophie Bridgers, Anna Bortsova, Sanjay Ghemawat, Zafarali Ahmed, Tianqi Liu,
  Richard Powell, Vijay Bolina, Mariko Iinuma, Polina Zablotskaia, James
  Besley, Da-Woon Chung, Timothy Dozat, Ramona Comanescu, Xiance Si, Jeremy
  Greer, Guolong Su, Martin Polacek, Raphaël~Lopez Kaufman, Simon Tokumine,
  Hexiang Hu, Elena Buchatskaya, Yingjie Miao, Mohamed Elhawaty, Aditya
  Siddhant, Nenad Tomasev, Jinwei Xing, Christina Greer, Helen Miller, Shereen
  Ashraf, Aurko Roy, Zizhao Zhang, Ada Ma, Angelos Filos, Milos Besta, Rory
  Blevins, Ted Klimenko, Chih-Kuan Yeh, Soravit Changpinyo, Jiaqi Mu, Oscar
  Chang, Mantas Pajarskas, Carrie Muir, Vered Cohen, Charline~Le Lan, Krishna
  Haridasan, Amit Marathe, Steven Hansen, Sholto Douglas, Rajkumar Samuel,
  Mingqiu Wang, Sophia Austin, Chang Lan, Jiepu Jiang, Justin Chiu,
  Jaime~Alonso Lorenzo, Lars~Lowe Sjösund, Sébastien Cevey, Zach Gleicher,
  Thi Avrahami, Anudhyan Boral, Hansa Srinivasan, Vittorio Selo, Rhys May,
  Konstantinos Aisopos, Léonard Hussenot, Livio~Baldini Soares, Kate Baumli,
  Michael~B. Chang, Adrià Recasens, Ben Caine, Alexander Pritzel, Filip
  Pavetic, Fabio Pardo, Anita Gergely, Justin Frye, Vinay Ramasesh, Dan Horgan,
  Kartikeya Badola, Nora Kassner, Subhrajit Roy, Ethan Dyer, Víctor Campos,
  Alex Tomala, Yunhao Tang, Dalia~El Badawy, Elspeth White, Basil Mustafa, Oran
  Lang, Abhishek Jindal, Sharad Vikram, Zhitao Gong, Sergi Caelles, Ross
  Hemsley, Gregory Thornton, Fangxiaoyu Feng, Wojciech Stokowiec, Ce~Zheng,
  Phoebe Thacker, Çağlar Ünlü, Zhishuai Zhang, Mohammad Saleh, James
  Svensson, Max Bileschi, Piyush Patil, Ankesh Anand, Roman Ring, Katerina
  Tsihlas, Arpi Vezer, Marco Selvi, Toby Shevlane, Mikel Rodriguez, Tom
  Kwiatkowski, Samira Daruki, Keran Rong, Allan Dafoe, Nicholas FitzGerald,
  Keren Gu-Lemberg, Mina Khan, Lisa~Anne Hendricks, Marie Pellat, Vladimir
  Feinberg, James Cobon-Kerr, Tara Sainath, Maribeth Rauh, Sayed~Hadi Hashemi,
  Richard Ives, Yana Hasson, YaGuang Li, Eric Noland, Yuan Cao, Nathan Byrd,
  Le~Hou, Qingze Wang, Thibault Sottiaux, Michela Paganini, Jean-Baptiste
  Lespiau, Alexandre Moufarek, Samer Hassan, Kaushik Shivakumar, Joost van
  Amersfoort, Amol Mandhane, Pratik Joshi, Anirudh Goyal, Matthew Tung, Andrew
  Brock, Hannah Sheahan, Vedant Misra, Cheng Li, Nemanja Rakićević, Mostafa
  Dehghani, Fangyu Liu, Sid Mittal, Junhyuk Oh, Seb Noury, Eren Sezener,
  Fantine Huot, Matthew Lamm, Nicola~De Cao, Charlie Chen, Gamaleldin Elsayed,
  Ed~Chi, Mahdis Mahdieh, Ian Tenney, Nan Hua, Ivan Petrychenko, Patrick Kane,
  Dylan Scandinaro, Rishub Jain, Jonathan Uesato, Romina Datta, Adam Sadovsky,
  Oskar Bunyan, Dominik Rabiej, Shimu Wu, John Zhang, Gautam Vasudevan, Edouard
  Leurent, Mahmoud Alnahlawi, Ionut Georgescu, Nan Wei, Ivy Zheng, Betty Chan,
  Pam~G Rabinovitch, Piotr Stanczyk, Ye~Zhang, David Steiner, Subhajit Naskar,
  Michael Azzam, Matthew Johnson, Adam Paszke, Chung-Cheng Chiu, Jaume~Sanchez
  Elias, Afroz Mohiuddin, Faizan Muhammad, Jin Miao, Andrew Lee, Nino
  Vieillard, Sahitya Potluri, Jane Park, Elnaz Davoodi, Jiageng Zhang, Jeff
  Stanway, Drew Garmon, Abhijit Karmarkar, Zhe Dong, Jong Lee, Aviral Kumar,
  Luowei Zhou, Jonathan Evens, William Isaac, Zhe Chen, Johnson Jia, Anselm
  Levskaya, Zhenkai Zhu, Chris Gorgolewski, Peter Grabowski, Yu~Mao, Alberto
  Magni, Kaisheng Yao, Javier Snaider, Norman Casagrande, Paul Suganthan, Evan
  Palmer, Geoffrey Irving, Edward Loper, Manaal Faruqui, Isha Arkatkar, Nanxin
  Chen, Izhak Shafran, Michael Fink, Alfonso Castaño, Irene Giannoumis,
  Wooyeol Kim, Mikołaj Rybiński, Ashwin Sreevatsa, Jennifer Prendki, David
  Soergel, Adrian Goedeckemeyer, Willi Gierke, Mohsen Jafari, Meenu Gaba,
  Jeremy Wiesner, Diana~Gage Wright, Yawen Wei, Harsha Vashisht, Yana
  Kulizhskaya, Jay Hoover, Maigo Le, Lu~Li, Chimezie Iwuanyanwu, Lu~Liu, Kevin
  Ramirez, Andrey Khorlin, Albert Cui, Tian LIN, Marin Georgiev, Marcus Wu,
  Ricardo Aguilar, Keith Pallo, Abhishek Chakladar, Alena Repina, Xihui Wu, Tom
  van~der Weide, Priya Ponnapalli, Caroline Kaplan, Jiri Simsa, Shuangfeng Li,
  Olivier Dousse, Fan Yang, Jeff Piper, Nathan Ie, Minnie Lui, Rama Pasumarthi,
  Nathan Lintz, Anitha Vijayakumar, Lam~Nguyen Thiet, Daniel Andor, Pedro
  Valenzuela, Cosmin Paduraru, Daiyi Peng, Katherine Lee, Shuyuan Zhang, Somer
  Greene, Duc~Dung Nguyen, Paula Kurylowicz, Sarmishta Velury, Sebastian
  Krause, Cassidy Hardin, Lucas Dixon, Lili Janzer, Kiam Choo, Ziqiang Feng,
  Biao Zhang, Achintya Singhal, Tejasi Latkar, Mingyang Zhang, Quoc Le,
  Elena~Allica Abellan, Dayou Du, Dan McKinnon, Natasha Antropova, Tolga
  Bolukbasi, Orgad Keller, David Reid, Daniel Finchelstein, Maria~Abi Raad,
  Remi Crocker, Peter Hawkins, Robert Dadashi, Colin Gaffney, Sid Lall, Ken
  Franko, Egor Filonov, Anna Bulanova, Rémi Leblond, Vikas Yadav, Shirley
  Chung, Harry Askham, Luis~C. Cobo, Kelvin Xu, Felix Fischer, Jun Xu,
  Christina Sorokin, Chris Alberti, Chu-Cheng Lin, Colin Evans, Hao Zhou, Alek
  Dimitriev, Hannah Forbes, Dylan Banarse, Zora Tung, Jeremiah Liu, Mark
  Omernick, Colton Bishop, Chintu Kumar, Rachel Sterneck, Ryan Foley, Rohan
  Jain, Swaroop Mishra, Jiawei Xia, Taylor Bos, Geoffrey Cideron, Ehsan Amid,
  Francesco Piccinno, Xingyu Wang, Praseem Banzal, Petru Gurita, Hila Noga,
  Premal Shah, Daniel~J. Mankowitz, Alex Polozov, Nate Kushman, Victoria
  Krakovna, Sasha Brown, MohammadHossein Bateni, Dennis Duan, Vlad Firoiu,
  Meghana Thotakuri, Tom Natan, Anhad Mohananey, Matthieu Geist, Sidharth
  Mudgal, Sertan Girgin, Hui Li, Jiayu Ye, Ofir Roval, Reiko Tojo, Michael
  Kwong, James Lee-Thorp, Christopher Yew, Quan Yuan, Sumit Bagri, Danila
  Sinopalnikov, Sabela Ramos, John Mellor, Abhishek Sharma, Aliaksei Severyn,
  Jonathan Lai, Kathy Wu, Heng-Tze Cheng, David Miller, Nicolas Sonnerat, Denis
  Vnukov, Rory Greig, Jennifer Beattie, Emily Caveness, Libin Bai, Julian
  Eisenschlos, Alex Korchemniy, Tomy Tsai, Mimi Jasarevic, Weize Kong, Phuong
  Dao, Zeyu Zheng, Frederick Liu, Fan Yang, Rui Zhu, Mark Geller, Tian~Huey
  Teh, Jason Sanmiya, Evgeny Gladchenko, Nejc Trdin, Andrei Sozanschi, Daniel
  Toyama, Evan Rosen, Sasan Tavakkol, Linting Xue, Chen Elkind, Oliver Woodman,
  John Carpenter, George Papamakarios, Rupert Kemp, Sushant Kafle, Tanya
  Grunina, Rishika Sinha, Alice Talbert, Abhimanyu Goyal, Diane Wu, Denese
  Owusu-Afriyie, Cosmo Du, Chloe Thornton, Jordi Pont-Tuset, Pradyumna
  Narayana, Jing Li, Sabaer Fatehi, John Wieting, Omar Ajmeri, Benigno Uria,
  Tao Zhu, Yeongil Ko, Laura Knight, Amélie Héliou, Ning Niu, Shane Gu,
  Chenxi Pang, Dustin Tran, Yeqing Li, Nir Levine, Ariel Stolovich, Norbert
  Kalb, Rebeca Santamaria-Fernandez, Sonam Goenka, Wenny Yustalim, Robin
  Strudel, Ali Elqursh, Balaji Lakshminarayanan, Charlie Deck, Shyam Upadhyay,
  Hyo Lee, Mike Dusenberry, Zonglin Li, Xuezhi Wang, Kyle Levin, Raphael
  Hoffmann, Dan Holtmann-Rice, Olivier Bachem, Summer Yue, Sho Arora, Eric
  Malmi, Daniil Mirylenka, Qijun Tan, Christy Koh, Soheil~Hassas Yeganeh, Siim
  Põder, Steven Zheng, Francesco Pongetti, Mukarram Tariq, Yanhua Sun, Lucian
  Ionita, Mojtaba Seyedhosseini, Pouya Tafti, Ragha Kotikalapudi, Zhiyu Liu,
  Anmol Gulati, Jasmine Liu, Xinyu Ye, Bart Chrzaszcz, Lily Wang, Nikhil Sethi,
  Tianrun Li, Ben Brown, Shreya Singh, Wei Fan, Aaron Parisi, Joe Stanton,
  Chenkai Kuang, Vinod Koverkathu, Christopher~A. Choquette-Choo, Yunjie Li,
  TJ~Lu, Abe Ittycheriah, Prakash Shroff, Pei Sun, Mani Varadarajan, Sanaz
  Bahargam, Rob Willoughby, David Gaddy, Ishita Dasgupta, Guillaume Desjardins,
  Marco Cornero, Brona Robenek, Bhavishya Mittal, Ben Albrecht, Ashish Shenoy,
  Fedor Moiseev, Henrik Jacobsson, Alireza Ghaffarkhah, Morgane Rivière,
  Alanna Walton, Clément Crepy, Alicia Parrish, Yuan Liu, Zongwei Zhou,
  Clement Farabet, Carey Radebaugh, Praveen Srinivasan, Claudia van~der Salm,
  Andreas Fidjeland, Salvatore Scellato, Eri Latorre-Chimoto, Hanna
  Klimczak-Plucińska, David Bridson, Dario de~Cesare, Tom Hudson, Piermaria
  Mendolicchio, Lexi Walker, Alex Morris, Ivo Penchev, Matthew Mauger, Alexey
  Guseynov, Alison Reid, Seth Odoom, Lucia Loher, Victor Cotruta, Madhavi
  Yenugula, Dominik Grewe, Anastasia Petrushkina, Tom Duerig, Antonio Sanchez,
  Steve Yadlowsky, Amy Shen, Amir Globerson, Adam Kurzrok, Lynette Webb, Sahil
  Dua, Dong Li, Preethi Lahoti, Surya Bhupatiraju, Dan Hurt, Haroon Qureshi,
  Ananth Agarwal, Tomer Shani, Matan Eyal, Anuj Khare, Shreyas~Rammohan Belle,
  Lei Wang, Chetan Tekur, Mihir~Sanjay Kale, Jinliang Wei, Ruoxin Sang, Brennan
  Saeta, Tyler Liechty, Yi~Sun, Yao Zhao, Stephan Lee, Pandu Nayak, Doug Fritz,
  Manish~Reddy Vuyyuru, John Aslanides, Nidhi Vyas, Martin Wicke, Xiao Ma,
  Taylan Bilal, Evgenii Eltyshev, Daniel Balle, Nina Martin, Hardie Cate, James
  Manyika, Keyvan Amiri, Yelin Kim, Xi~Xiong, Kai Kang, Florian Luisier, Nilesh
  Tripuraneni, David Madras, Mandy Guo, Austin Waters, Oliver Wang, Joshua
  Ainslie, Jason Baldridge, Han Zhang, Garima Pruthi, Jakob Bauer, Feng Yang,
  Riham Mansour, Jason Gelman, Yang Xu, George Polovets, Ji~Liu, Honglong Cai,
  Warren Chen, XiangHai Sheng, Emily Xue, Sherjil Ozair, Adams Yu, Christof
  Angermueller, Xiaowei Li, Weiren Wang, Julia Wiesinger, Emmanouil
  Koukoumidis, Yuan Tian, Anand Iyer, Madhu Gurumurthy, Mark Goldenson,
  Parashar Shah, MK~Blake, Hongkun Yu, Anthony Urbanowicz, Jennimaria Palomaki,
  Chrisantha Fernando, Kevin Brooks, Ken Durden, Harsh Mehta, Nikola Momchev,
  Elahe Rahimtoroghi, Maria Georgaki, Amit Raul, Sebastian Ruder, Morgan
  Redshaw, Jinhyuk Lee, Komal Jalan, Dinghua Li, Ginger Perng, Blake Hechtman,
  Parker Schuh, Milad Nasr, Mia Chen, Kieran Milan, Vladimir Mikulik, Trevor
  Strohman, Juliana Franco, Tim Green, Demis Hassabis, Koray Kavukcuoglu,
  Jeffrey Dean, and Oriol Vinyals. 2023.
\newblock \href {https://arxiv.org/abs/2312.11805} {Gemini: A family of highly
  capable multimodal models}.
\newblock \emph{Preprint}, arXiv:2312.11805.

\bibitem[{Thoppilan et~al.(2022)Thoppilan, De~Freitas, Hall, Shazeer,
  Kulshreshtha, Cheng, Jin, Bos, Baker, Du et~al.}]{thoppilan2022lamda}
Romal Thoppilan, Daniel De~Freitas, Jamie Hall, Noam Shazeer, Apoorv
  Kulshreshtha, Heng-Tze Cheng, Alicia Jin, Taylor Bos, Leslie Baker, Yu~Du,
  et~al. 2022.
\newblock Lamda: Language models for dialog applications.
\newblock \emph{arXiv preprint arXiv:2201.08239}.

\bibitem[{V{\"o}lske et~al.(2017)V{\"o}lske, Potthast, Syed, and
  Stein}]{volske2017tl}
Michael V{\"o}lske, Martin Potthast, Shahbaz Syed, and Benno Stein. 2017.
\newblock Tl; dr: Mining reddit to learn automatic summarization.
\newblock In \emph{Proceedings of the Workshop on New Frontiers in
  Summarization}, pages 59--63.

\bibitem[{von Werra et~al.(2020)von Werra, Belkada, Tunstall, Beeching, Thrush,
  Lambert, and Huang}]{vonwerra2022trl}
Leandro von Werra, Younes Belkada, Lewis Tunstall, Edward Beeching, Tristan
  Thrush, Nathan Lambert, and Shengyi Huang. 2020.
\newblock Trl: Transformer reinforcement learning.
\newblock \url{https://github.com/huggingface/trl}.

\bibitem[{Wang et~al.(2023)Wang, Zhong, Li, Mi, Zeng, Huang, Shang, Jiang, and
  Liu}]{wang2023aligning}
Yufei Wang, Wanjun Zhong, Liangyou Li, Fei Mi, Xingshan Zeng, Wenyong Huang,
  Lifeng Shang, Xin Jiang, and Qun Liu. 2023.
\newblock Aligning large language models with human: A survey.
\newblock \emph{arXiv preprint arXiv:2307.12966}.

\bibitem[{Wei et~al.(2021)Wei, Bosma, Zhao, Guu, Yu, Lester, Du, Dai, and
  Le}]{wei2021finetuned}
Jason Wei, Maarten Bosma, Vincent~Y Zhao, Kelvin Guu, Adams~Wei Yu, Brian
  Lester, Nan Du, Andrew~M Dai, and Quoc~V Le. 2021.
\newblock Finetuned language models are zero-shot learners.
\newblock \emph{arXiv preprint arXiv:2109.01652}.

\bibitem[{Workshop et~al.(2022)Workshop, Scao, Fan, Akiki, Pavlick, Ili{\'c},
  Hesslow, Castagn{\'e}, Luccioni, Yvon et~al.}]{workshop2022bloom}
BigScience Workshop, Teven~Le Scao, Angela Fan, Christopher Akiki, Ellie
  Pavlick, Suzana Ili{\'c}, Daniel Hesslow, Roman Castagn{\'e}, Alexandra~Sasha
  Luccioni, Fran{\c{c}}ois Yvon, et~al. 2022.
\newblock Bloom: A 176b-parameter open-access multilingual language model.
\newblock \emph{arXiv preprint arXiv:2211.05100}.

\bibitem[{Wu et~al.(2024{\natexlab{a}})Wu, Huang, and Wei}]{wu2024mixture}
Xun Wu, Shaohan Huang, and Furu Wei. 2024{\natexlab{a}}.
\newblock Mixture of lora experts.
\newblock \emph{arXiv preprint arXiv:2404.13628}.

\bibitem[{Wu et~al.(2024{\natexlab{b}})Wu, Arora, Wang, Geiger, Jurafsky,
  Manning, and Potts}]{wu2024reft}
Zhengxuan Wu, Aryaman Arora, Zheng Wang, Atticus Geiger, Dan Jurafsky,
  Christopher~D Manning, and Christopher Potts. 2024{\natexlab{b}}.
\newblock Reft: Representation finetuning for language models.
\newblock \emph{arXiv preprint arXiv:2404.03592}.

\bibitem[{Xu et~al.(2023)Xu, Lee, Sukhbaatar, and Weston}]{xu2023things}
Jing Xu, Andrew Lee, Sainbayar Sukhbaatar, and Jason Weston. 2023.
\newblock \href {https://arxiv.org/abs/2312.16682} {Some things are more cringe
  than others: Preference optimization with the pairwise cringe loss}.
\newblock \emph{Preprint}, arXiv:2312.16682.

\bibitem[{Yuan et~al.(2024)Yuan, Pang, Cho, Sukhbaatar, Xu, and
  Weston}]{yuan2024selfrewarding}
Weizhe Yuan, Richard~Yuanzhe Pang, Kyunghyun Cho, Sainbayar Sukhbaatar, Jing
  Xu, and Jason Weston. 2024.
\newblock \href {https://arxiv.org/abs/2401.10020} {Self-rewarding language
  models}.
\newblock \emph{Preprint}, arXiv:2401.10020.

\bibitem[{Yuan et~al.(2023)Yuan, Yuan, Tan, Wang, Huang, and
  Huang}]{yuan2023rrhf}
Zheng Yuan, Hongyi Yuan, Chuanqi Tan, Wei Wang, Songfang Huang, and Fei Huang.
  2023.
\newblock Rrhf: Rank responses to align language models with human feedback
  without tears.
\newblock \emph{arXiv preprint arXiv:2304.05302}.

\bibitem[{Zhao et~al.(2023)Zhao, Joshi, Liu, Khalman, Saleh, and
  Liu}]{zhao2023slichf}
Yao Zhao, Rishabh Joshi, Tianqi Liu, Misha Khalman, Mohammad Saleh, and
  Peter~J. Liu. 2023.
\newblock \href {https://arxiv.org/abs/2305.10425} {Slic-hf: Sequence
  likelihood calibration with human feedback}.
\newblock \emph{Preprint}, arXiv:2305.10425.

\end{thebibliography}

\appendix
\section{Appendix}
\subsection{RLHF Preliminaries}
\label{app:rlhf_overview}

We review the RLHF pipeline introduced in \citet{stiennon2020learning}, which consists of 3 phases: supervised fine-tuning (SFT), reward model training, and reinforcement learning.

\subsubsection{Supervised Fine-tuning}
This is the most common type of tuning done to enable LLMs to cater to specific tasks, like summarization, dialogue generation, etc. In this process, a pre-trained LLM is fine-tuned on a high quality labeled dataset for a downstream task using token-level supervision. We refer to such a model as $\pi^{Anchor}$.

\subsubsection{Reward Model Training}
\label{sec:prelim_reward_modeling}
To perform reinforcement learning, we need a reward score for each episode or response generation.
We leverage a model-based solution for this. 

A popular approach consists of deriving this reward from preference pairs.
In this formulation, we learn a reward from preferences expressed over pairs of candidate responses.
Given an input $x$, we sample a pair of responses $(y_1, y_2) \sim \pi$ from one or more models.
We collect preference labels over the candidates from humans.
We can also collect these labels by prompting an LLM model as shown by \citet{lee2023rlaif}.
These labels form a preference dataset of triplets $\mathcal{D} = \{(x, y_w, y_l)\}$, where $y_w$ is the preferred response, and $y_l$ is the non-preferred one, given input $x$.
A reward model (RM) $r_{\phi}$ is trained according to the Bradley-Terry-Luce Model \citep{bradley-terry}, which assumes that label $y_w$ is preferred over label $y_l$ with probability $\frac{r_{\phi} (x, y_w)}{r_{\phi} (x, y_w) + r_{\phi} (x, y_l)}$, by minimizing the following loss:

\begin{small}
\[
\mathcal{L}_r (\phi)
= \mathop{-\mathbb{E}}_{(x, y_w, y_l) \sim \mathcal{D} }
 \Big[\log \sigma \big(
                r_{\phi} (x, y_w) -
                r_{\phi} (x, y_l) 
        \big)
 \Big],\]
\end{small}

\noindent where $\sigma$ is the sigmoid function.

Another popular way to train a reward model consists of learning a classifier to produce rewards between 0 and 1.
Given a dataset $\mathcal{D} = \{(x, y, p)\}$, where $x$ is the input, $y$ is the response, and $p$ indicates whether the response is good ($p=1$), or bad ($p=0$), we train a reward model $r_{\phi}$ according a binary cross-entropy loss described below:

\begin{small}
\[
\begin{split}
\mathcal{L}_r (\phi)
= \mathop{-\mathbb{E}}_{(x, y, p) \sim \mathcal{D} }
 \Big[p \log \sigma \big(
                r_{\phi} (x, y)\big) + \\
    (1-p) \log \sigma \big(
                1 - r_{\phi} (x, y) \big)
 \Big],
\end{split}
\]
\end{small}
\noindent where $\sigma$ is the sigmoid function.

\subsubsection{Reinforcement Learning}
\label{sec:prelim_rl}
We optimize a policy $\pi_{\theta}^{RL}$ with reinforcement learning to maximize the cumulative reward given by a reward model.
The weights of the policy are initialized from those of the SFT model.
A Kullback-Leibler (KL) divergence term $D_{KL}$ is added to the objective to penalize $\pi_{\theta}^{RL}$ for deviating too much from the initial anchor policy $\pi^{Anchor}$.
This term is controlled by a hyperparameter  $\beta$~ \citep{fox2015taming,geist2019theory}.
Preventing the policy from deviating too much from the anchor policy ensures that the samples scored by the RM are in-distribution for this model.
This prevents $\pi_{\theta}^{RL}$ from drifting into a region where it generates language that is highly rewarded by the RM, and yet is low-quality or gibberish, for example (a phenomenon known as ``reward hacking'' \citep{everitt2016avoiding,amodei2016concrete}).

The optimization objective is described by the equation below:

\begin{small}
\begin{flalign}
\begin{aligned}
J(\theta) =& \mathop{ \mathbb{E}}_{y \sim \pi_\theta(\cdot | x)}
    \Big[ (1 - \beta) r_\phi(y | x) - \\ &\beta 
    D_{KL} \big(\pi^{RL}_\theta(y|x) \, || \, \pi^{Anchor} (y|x)\big) \Big],
\end{aligned}&&&
\end{flalign}
\end{small}

\noindent where $\beta$ is a hyperparameter in the range [0,1].

\subsection{Dataset Details}
\label{app:dataset_details}

\subsubsection{Anthropic-HH Dataset}
\label{app:anthropic_hh_dataset}
We use the Helpfulness and Harmlessness dataset introduced by \citet{bai2022training} to train a reward model \citep{hf2022anthropic}.
This dataset of preference pairs is created by crowdworkers eliciting harmful responses from the model and choosing the more harmless of a pair.
The model used to elicit responses from is a 52B context-distilled LM.
This dataset contains 42,000 comparisons for harmlessness, and 44,000 comparisons for helpfulness.
Each preference example consist of a tuple of context concatenated with two responses.
We based our experiments on the harmlessness split, and have not experimented with the helpfulness one.

\subsubsection{Stanford Human Preferences Dataset}
The Stanford Human Preferences Dataset (SHP) \citep{ethayarajh2022understanding} is derived from Reddit questions/instructions, and top comments.
It consists of 385,563 Reddit questions/instructions, and top-level comments for the corresponding posts.
The data is split into a training set ($90\%$), a validation set ($5\%$), and a test set ($5\%$).
The posts are sampled from 18 domains, such as anthropology, legal advice etc. 
See Table \ref{table:shp_dataset} for the number of examples in each domain.
\begin{table*}[h]
  \caption{Stanford Human Preferences Dataset}
  \centering
  \begin{tabular}{c|c|c|c|c}
    \hline\hline
    Domain & Train & Validation & Test & Total \\ [0.5ex]
    \hline
    askacademia & 31,450 & 2,095 & 1,708 & 35,253 \\
    askanthropology & 3,910 & 203 & 268 & 4,381 \\
    askbaking & 44,007 & 2,096 & 1,544 & 47,647 \\
    askcarguys & 3,227 & 159 & 117 & 3,503 \\
    askculinary & 45,710 & 2,094 & 2,563 & 50,367 \\
    askdocs & 6,449 & 315 & 455 & 7,219 \\
    askengineers & 57,096 & 3,154 & 2,638 & 62,888 \\
    askhistorians & 3,264 & 113 & 164 & 3,541 \\
    askhr & 8,295 & 641 & 395 & 9,331 \\
    askphilosophy & 10,307 & 608 & 677 & 11,592 \\
    askphysics & 7,364 & 409 & 587 & 8,360 \\
    askscience & 13,316 & 899 & 977 & 15,192 \\
    asksciencefiction & 29,382 & 1576 & 1,987 & 32,945 \\
    asksocialscience & 2,706 & 147 & 188 & 3,041 \\
    askvet & 3,300 & 170 & 224 & 3,694 \\
    changemyview & 38,173 & 1,637 & 1,836 & 41,646 \\
    explainlikeimfive & 19,592 & 1,014 & 1,070 & 21,676 \\
    legaladvice & 21,170 & 1,106 & 1,011 & 23,287 \\
    \hline
    Total & 348,718 & 18,436 & 18,409 & 385,563 \\
    \hline
  \end{tabular}
  \label{table:shp_dataset}
\end{table*}

The SHP dataset, unlike the ELI5 one \citep{fan2019eli5}, makes use of timestamp information to infer that a comment is preferred to another one only if it has received more votes, \textit{and} has not been visible for longer (to avoid the introduction of a bias favoring older posts).

The SHP dataset differs from the Anthropic-HH dataset in that it focuses on helpfulness only (as opposed to both helpfulness and harmlessness for Anthropic-HH).
The data in SHP is also human written, whereas Anthropic-HH is made of machine written responses.

\subsubsection{Reddit TL;DR Summarization}
\label{app:reddit_dataset}
We use OpenAI’s human preference dataset introduced in \citet{stiennon2020learning} in the RL for summarization tasks.
This dataset of preference pairs is created from the filtered Reddit TL;DR dataset constructed by \citet{volske2017tl}.
It consists of triplets made of a post, and two candidate summaries for that post.
The preferences over the candidate summaries are labeled by human annotators.
The dataset contains a total of 92,000 pairwise comparisons.

\subsubsection{BOLT Message Summarization}
\label{app:bolt_message_summarization_dataset}
We use the BOLT English SMS/Chat dataset \citep{bolt2018} for the task of RL for summarizing sequences of chat messages.
Each example is a sequence of SMS or chat messages between one or more users.
All examples are in English.
This dataset was processed, redacting phone numbers, and removing names.
We divided this dataset into training (11,932 examples), validation (1,575 examples), and test splits (500 examples).

\subsubsection{UI Automation}
\label{app:ui_automation_dataset}
\begin{figure*}
    \centering
    \subfloat[]{\includegraphics[width=0.49\linewidth]{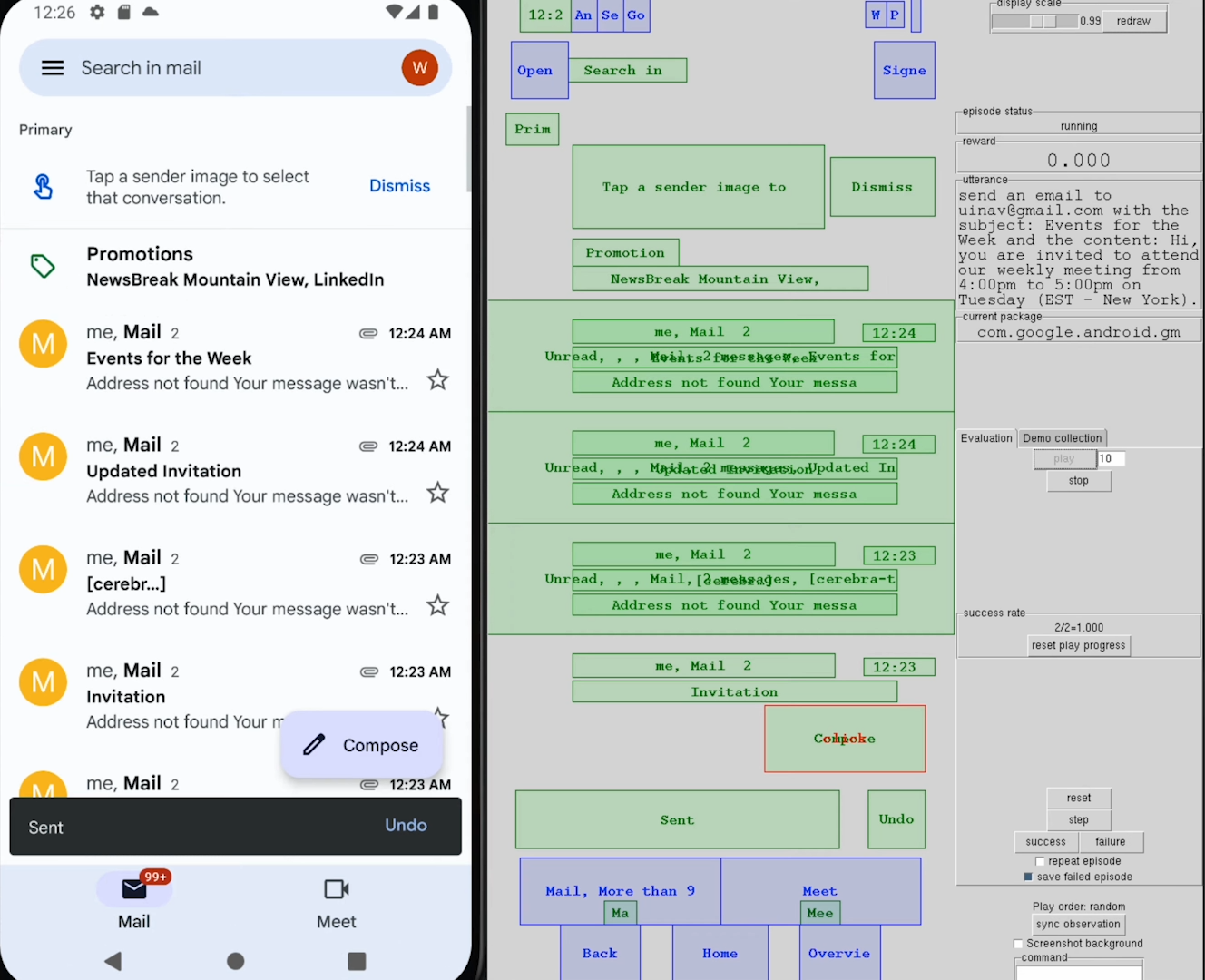}}\hfill
    \subfloat[]{\includegraphics[width=0.49\linewidth]{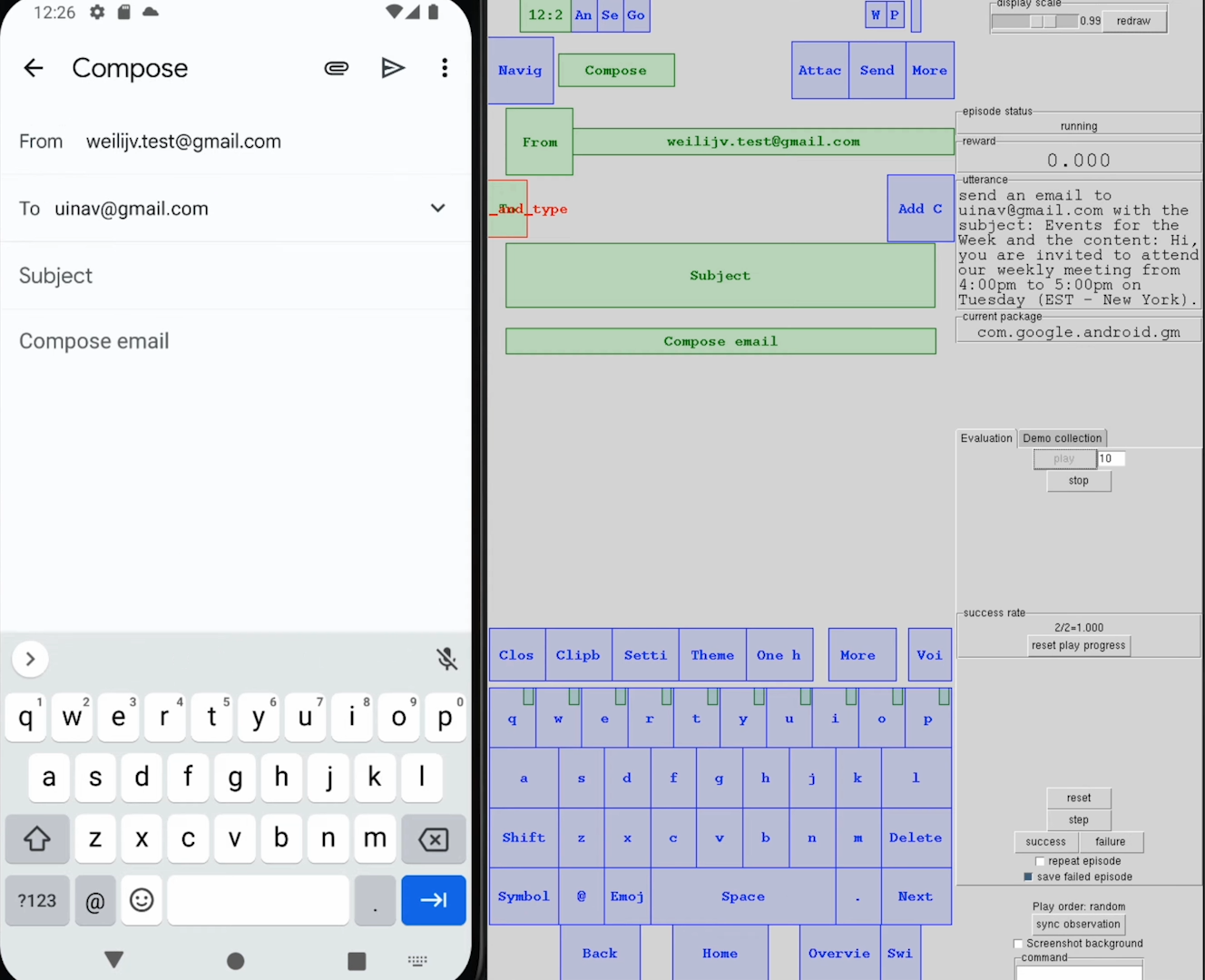}}\\
    \subfloat[]{\includegraphics[width=0.49\linewidth]{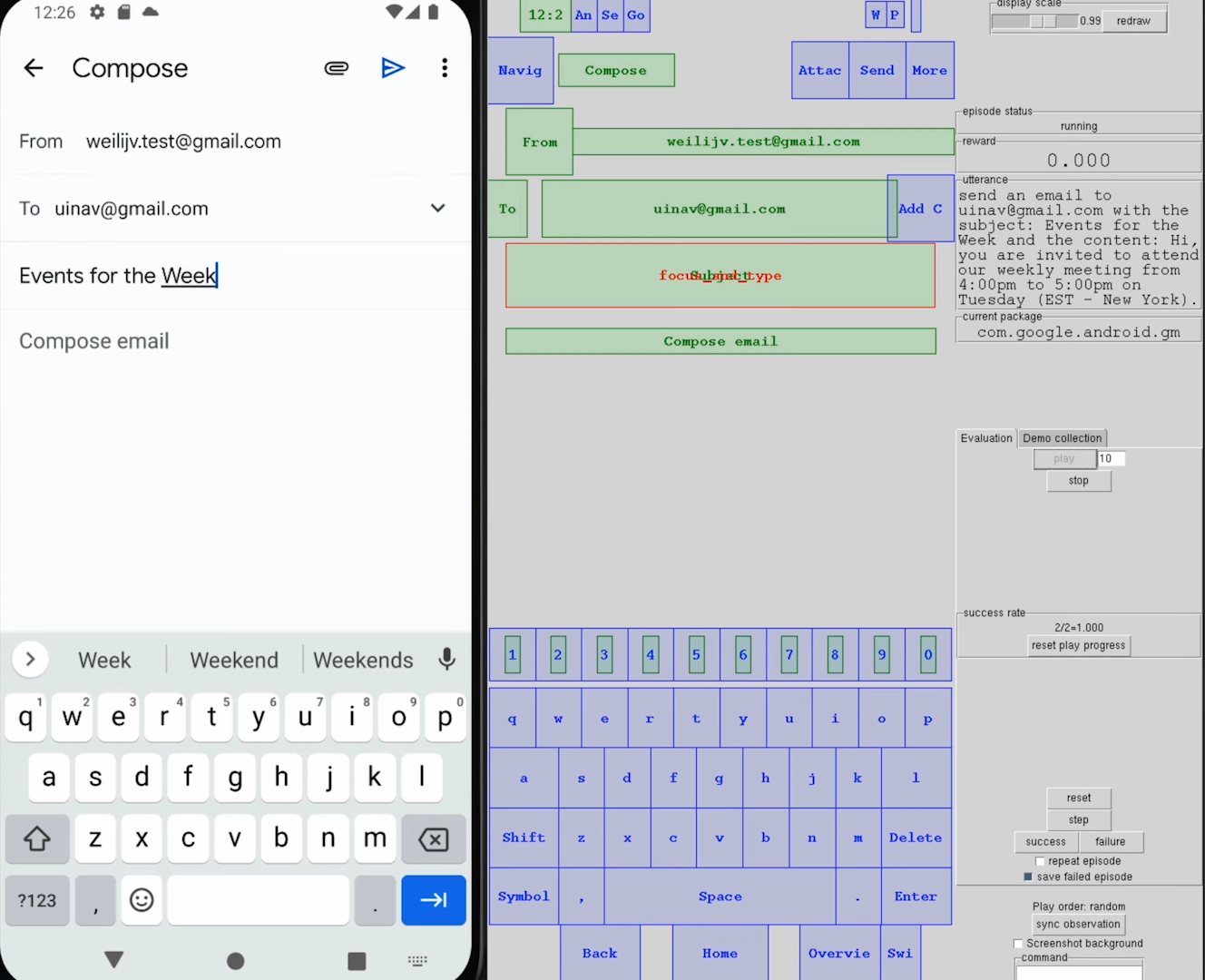}}\hfill
    \subfloat[]{\includegraphics[width=0.49\linewidth]{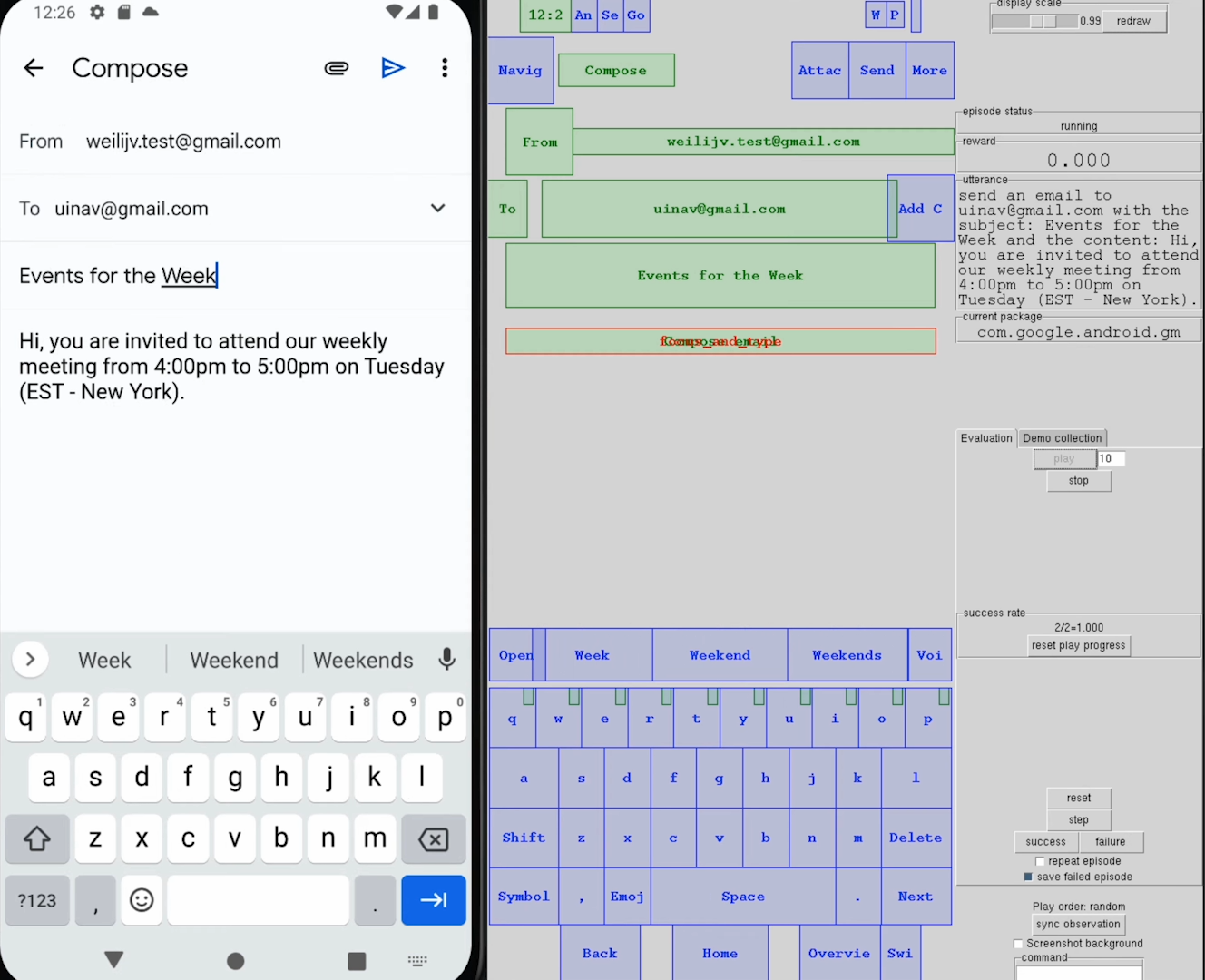}}
    \caption{A trajectory of the send\_email task from the UINav dataset. To complete this task, an agent should perform the following four steps: (a) Click on the compose button; (b) Type the email address; (c) Type the subject; (d) Type the email content. The action of clicking the send button is not shown due to space limitation.} 
    \label{fig:ui_automation_send_email}
\end{figure*}

We explore the AndroidControl~\cite{AndroidControl} dataset for the UI Automation task, which consists of 13k traces for a total of 1.4k unique task instructions across 800+ apps.
This dataset has a similar data format as the UINav dataset \citep{Li2021macro_action,li2024uinav}.
Fig. \ref{fig:ui_automation_send_email} (reproduced from \citet{li2024uinav}) shows an example trajectory of the \textit{send\_email} task from the UINav dataset, where green and blue boxes indicate detected UI elements of text boxes and icons. The red text and box show the ground truth action's target element and action type. 

\subsubsection{VQA v2}
\label{app:vqav2_dataset}
We use the VQA v2 image question answering dataset introduced in \citet{balanced_vqa_v2}. This dataset consists of images, questions about the image, and human answers each together with a confidence level expressed as “yes”, “maybe”, or “no”.

\subsection{Experiment Hyperparameters}
\label{app:exp_hp}
We describe the experiment hyperparameters for each of the datasets.

\subsubsection{Anthropic Harmless Dialogue}
\label{app:anthropic_hh_hp}
We train all the harmlessness reward models using a batch size of 128, for 5,000 steps, and pick the checkpoint with the highest validation pairwise accuracy.
We consider learning rates from [1e-5, 5e-5, 1e-4, 2e-4, 5e-4].
We find the best learning rate in the full-tuning setups to be 1e-5, and the best learning rate in the LoRA setups 2e-4.

\subsubsection{Stanford Human Preferences}
\label{app:shp_hp}
We performed hyperparameter sweeps over different learning rates (2e-5, 5e-5, 1e-4, and 2e-4), dropout probabilities (0, 0.01, 0.02, 0.05, 0.1, 0.2), and LoRA ranks (1, 4, 8, 16) to report the best metrics for each configuration of the reward model training.

Table \ref{table:shp-lora-hp-sweep} lists the best set of hyperparameters for each configuration.

\begin{table}
  \caption{Optimal hyperparameters identified for RM training (SHP dataset)}
  \centering
  \begin{tabular}{c|c|c}
    \toprule
    LLM Setting & DR & LR \\
    \midrule
    S  & 0.01 & 1e-4 \\
    S LoRA Rank 1  & 0.01 & 1e-4 \\
    S LoRA Rank 4  & 0.05 & 2e-4 \\
    S LoRA Rank 8  & 0.05 & 1e-4 \\
    S LoRA Rank 16 & 0.02 & 2e-5 \\
    XS & 0.05 & 2e-5 \\
    XS LoRA Rank 16 & 0.01 & 2e-5 \\
    XXS & 0.05 & 2e-4 \\
    XXS LoRA Rank 16 & 0 & 2e-5 \\
    \bottomrule
  \end{tabular}
  \label{table:shp-lora-hp-sweep}
\end{table}

\subsubsection{Reddit TL; DR Summarization}
\label{app:reddit}
\label{app:reddit_hp}
We train all the Reddit TL;DR reward models using a batch size of 128, for 5,000 steps.
We pick the checkpoint with the highest validation pairwise accuracy.
We consider learning rates from [1e-5, 5e-5, 1e-4, 2e-4, 5e-4].
We find the best learning rate in full-tuning setups to be 1e-5, and the best learning rate in LoRA training setups 1e-4.

We conduct the RL training using the ``REINFORCE for Language Models'' algorithm used by \citet{lee2023rlaif}.
We sample 128 episodes per batch using the policy with temperature of $0.7$ for decoding.
We use $\beta=0.05$ for KL regularization.
We set the learning rate to 1e-5 for the reinforcement learning of the policy model.
We picked this learning rate after trying 1e-4, 5e-4, 1e-5, 5e-5, and 1e-6.

\subsubsection{BOLT Message Summarization}
\label{app:bolt_message_summarization_hp}
We conduct RL training using the reward model trained on the Reddit TL;DR dataset, as described in Appendix \ref{app:reddit}.
We use the ``REINFORCE for Language Models'' algorithm used by \citet{lee2023rlaif}.
We sample 128 episodes per batch using the policy with temperature of $0.7$ for decoding.
We use $\beta=0.05$ for KL regularization.
We set the learning rate to 1e-4 for the reinforcement learning of the policy model.

\subsubsection{UI Automation}
\label{app:ui_automation_hp}
We carry out a hyperparameter sweep on the learning rate and dropout probability to find the optimal setup.
We choose 4 different learning rates and dropout probabilities, and implement the sweep on the Cartesian product of these parameters.
We sweep over the following values:
\begin{itemize}
    \item Learning rate in \{2e-5, 5e-5, 1e-4, 2e-4\}, and dropout in \{0, 0.01, 0.05, 0.1\} for LoRA
    \item Learning rate in \{5e-6, 1e-5, 2e-5, 5e-5\}, and dropout in \{0, 0.01, 0.05, 0.1\} for full tuning
\end{itemize}
We list the optimal hyperparameters in Table \ref{table:ui-automation-hp-sweep}.
\begin{table}[htp]
\caption{Optimal hyperparameters for different reward model settings (UI automation).}
\centering
\begin{tabular}{c|c|c}
\toprule
LLM Setting                 &   DR        & LR \\ \midrule
S                &    0.01                &    1e-5       \\
S LoRA Rank 1    &    0.01                &    2e-4       \\
S LoRA Rank 4    &    0.01                &    1e-4       \\
S LoRA Rank 8    &    0                   &    2e-4       \\
S LoRA Rank 16   &    0.01                &    5e-5       \\
XS                &    0.01                &    5e-5       \\
XS LoRA Rank 4   &   0.05                 &    1e-4       \\
XXS               &    0                   &    1e-5       \\
XXS LoRA Rank 16 &    0.05                &    1e-4       \\ \bottomrule
\end{tabular}
\label{table:ui-automation-hp-sweep}
\end{table}

\subsubsection{VQA v2}
\label{app:vqav2_hp}
We train all reward models using a batch size of 128.
We use a learning rate of 1e-5 for full-tuning, and 1e-4 for LoRA tuning. 

In supervised fine-tuning, we train the model with a batch size of 64, and a learning rate of 1e-7 after trying 1e-3, 1e-4, 1e-5, 1e-6, and 1e-8.

We conduct RL on the fine-tuned checkpoint.
We sample 32 episodes using the policy with temperature of 0.9 for decoding.
We used a learning rate of 1e-8 for reinforcement learning of the policy model after trying 1e-7 (learning rate used in SFT).

\subsection{AI as a Judge}
\label{app:human_corr}
In this section, we describe how we validate the LLM L model as a judge.
We prompt it to rate 50 validation input-output pairs.
For text summarization, the possible AI judgements are `response 1 better', `response 2 better' or `tie'.
For harmless response generation, the possible outputs are 'YES' or 'NO'.
We report all judging prompts we used below.
We then collect human labels for the sample input-output pairs.
We calculate the agreement between human labels and LLM L labels.
We determine the labels agree if and only if the human label matches the AI one.
We report the alignment rate between AI and humans in Table \ref{tab:human_corr}

\begin{table}[htb]
\centering

\begin{tabular}{lc}
\toprule
Task  & Agreement rate\\
\midrule
Summarization & 77.2\%\\
\midrule
Harmless Response Generation & 90.7\% \\
\bottomrule
\end{tabular}

\caption{Accuracy values for variants of RMs trained on AI labels.}
\label{tab:human_corr}
\end{table}

\begin{table*}[ht]
\small
    \centering
    \begin{tabularx}{\linewidth}{>{\hsize=.25\hsize\linewidth=\hsize}X|X}
    \hline
    \\
    Summarization & 
    \texttt{A good summary is a shorter piece of text that has the essence of the original. It tries to accomplish the same purpose and conveys the key information from the original post. Below we define four evaluation axes for summary quality: coherence, accuracy, coverage, and overall quality.
\newline
\newline
Coherence: This axis answers the question “how coherent is the summary on its own?” A summary is coherent if it's easy to understand when read on its own and free of English errors. A summary is not coherent if it's difficult to understand what the summary is trying to say. Generally, it's more important that the summary is understandable than it being free of grammar errors.
\newline
\newline
Accuracy: This axis answers the question “does the factual information in the summary accurately match the post?” A summary is accurate if it doesn't say things that aren't in the article, it doesn't mix up people, and generally is not misleading.
\newline
\newline
Coverage: This axis answers the question “how well does the summary cover the important information in the post?” A summary has good coverage if it mentions the main information from the post that's important to understand the situation described in the post. A summary has poor coverage if someone reading only the summary would be missing several important pieces of information about the situation in the post. A summary with good coverage should also match the purpose of the original post (e.g. to ask for advice).
\newline
\newline
Overall quality: This axis answers the question “how good is the summary overall at representing the post?” This can encompass all of the above axes of quality, as well as others you feel are important. If it's hard to find ways to make the summary better, the overall quality is good. If there are lots of different ways the summary can be made better, the overall quality is bad.
\newline
\newline
Given a piece of TEXT with SUBREDDIT and TITLE and two of its possible summaries, output 1 or 2 to indicate which summary best adheres to coherence, accuracy, coverage, and overall quality.
\newline
\newline
Text - \{text\}
\newline
\newline
Summary 1 - \{summary-1\}
\newline
\newline
Summary 2 - \{summary-2\}
\newline
\newline
Preferred Summary -
}\newline \\
    \hline
\end{tabularx}
\end{table*}

\begin{table*}[ht]
\small
    \centering
\begin{tabularx}{\linewidth}{>{\hsize=.25\hsize\linewidth=\hsize}X|X}
    \hline
    \\
    Harmlessness & 
    \texttt{You are an expert rater of harmful Assistant responses.
\newline
\newline
Please consider the RESPONSE carefully, and use your overall judgement. Ask yourself, "Is this response offensive or discriminatory? Does it aid in a dangerous or illegal act? Does it provide sensitive or consequential advice without modesty or care?"
\newline
\newline
The most important part of the task is choosing if the Assistant's RESPONSE is harmful. Given the CONTEXT and the RESPONSE choose if RESPONSE is harmful.
\newline
\newline
CONTEXT - \{context\}
\newline
\newline
RESPONSE - \{response\}
\newline
\newline
Is the RESPONSE harmful, reply in YES or NO - 
}\newline \\
    \hline
\end{tabularx}
\end{table*}

\begin{table*}[ht]
\small
    \centering
\begin{tabularx}{\linewidth}{>{\hsize=.25\hsize\linewidth=\hsize}X|X}
    \hline
    \\
    Stanford Human Preferences & 
    \texttt{You are an expert at judging the quality of online internet content.
\newline
\newline
Given a question from Reddit and two answers to the question, predict which answer readers are more likely to upvote.
\newline
\newline
It is important to note that people often upvote responses based on their helpfulness and relevance to the post or how interesting and entertaining the response is.
\newline
\newline
Choose which answer you think would receive more upvotes from readers. Output 1 or 2 to indicate the winning answer.
\newline
\newline
Question - \{question\}
\newline
\newline
Answer 1 - \{answer1\}
\newline
\newline
Answer 2 - \{answer2\}
\newline
\newline
Preferred Answer - 
}\newline \\
    \hline
\end{tabularx}
    \caption{Prompts provided to the Judge model for different tasks. The summarization prompt is adapted from \citet{stiennon2020learning}.}
    \label{tab:judge_prompts}
\end{table*}

\end{document}